\documentclass[sigconf,table]{acmart}

\renewcommand\footnotetextcopyrightpermission[1]{} 

\AtBeginDocument{%
  \providecommand\BibTeX{{%
    \normalfont B\kern-0.5em{\scshape i\kern-0.25em b}\kern-0.8em\TeX}}}

\usepackage{makecell} 
\usepackage[caption=false, listofformat=subsimple, labelformat=simple]{subfig} 

\usepackage[shortlabels]{enumitem}
\usepackage{multirow}
\usepackage{balance}
\usepackage{boxedminipage}
\usepackage{hyperref}
\usepackage{pdflscape}

\begin{document}

\title{EvoRobogami: Co-designing with Humans in Evolutionary Robotics Experiments}
\subtitle{with Supplementary Materials}
\subtitlenote{This paper is originally submitted to and accepted by GECCO 2022. The
           GECCO paper can be found at \url{https://doi.org/10.1145/3512290.3528867}.
           This arXiv paper contains supplementary figures and results that were not able
           to show up in the GECCO paper due to GECCO's page limit.}

\author{Huang Zonghao}
\affiliation{
  \institution{University of Pennsylvania}
  \streetaddress{220 S 33rd St}
  \city{Philadelphia}
  \state{Pennsylvania}
  \country{United States of America}}
\email{zoh@seas.upenn.edu}

\author{Quinn Wu}
\affiliation{
  \institution{University of Pennsylvania}
  \streetaddress{220 S 33rd St}
  \city{Philadelphia}
  \state{Pennsylvania}
  \country{United States of America}}
\email{quinnwu@seas.upenn.edu}

\author{David Howard}
\affiliation{
  \institution{CSIRO}
  \streetaddress{1 Technology Court Pullenvale}
  \city{Brisbane}
  \state{Queensland}
  \country{Australia}}
\email{david.howard@csiro.au}

\author{Cynthia Sung}
\affiliation{
  \institution{University of Pennsylvania}
  \streetaddress{220 S 33rd St}
  \city{Philadelphia}
  \state{Pennsylvania}
  \country{United States of America}}
\email{crsung@seas.upenn.edu}

\begin{abstract}
We study the effects of injecting human-generated designs into the initial
population of an evolutionary robotics experiment, where subsequent population
of robots are optimised via a Genetic Algorithm and MAP-Elites.  First, human
participants interact via a graphical front-end to explore a
directly-parameterised legged robot design space and attempt to produce robots
via a combination of intuition and trial-and-error that perform well in a range
of environments.  Environments are generated whose corresponding
high-performance robot designs range from intuitive to complex and hard to
grasp.  Once the human designs have been collected, their impact on the
evolutionary process is assessed by replacing a varying number of designs in the
initial population with human designs and subsequently running the evolutionary
algorithm.  Our results suggest that a balance of random and hand-designed
initial solutions provides the best performance for the problems considered, and
that human designs are most valuable when the problem is intuitive. The
influence of human design in an evolutionary algorithm is a highly understudied
area, and the insights in this paper may be valuable to the area of AI-based
design more generally.
\end{abstract}

\settopmatter{printacmref=false}\settopmatter{printacmref=false}
\maketitle
\pagestyle{plain} 

\section{Introduction}
\label{sec:introduction}
Evolutionary Robotics (ER) (e.g., \cite{bongard2013evolutionary}) is a powerful
tool for robot design, being able to explore interwoven design spaces of coupled
body, brain, and environmental interactions.  Its fitness-based performance
assessment is particularly useful in this role, as unintuitive, surprising
designs \cite{lehman2020surprising} can be assessed in a bias-free manner and
large, complex design spaces can be automatically explored in the pursuit of
desired behaviours.  ER is overwhelmingly implemented as a fully automated
process: a robot-producing black box that relies on computational power,
parallelisation, and extensive trial-and-error to tackle high-dimensional design
problems over a wide design space.  Conversely, human-centered design relies on
creativity, intuition, and domain knowledge that digital systems often struggle
to replicate, but typically a narrower design space.

\begin{figure}
    \centering
    \includegraphics[width=0.8\linewidth]{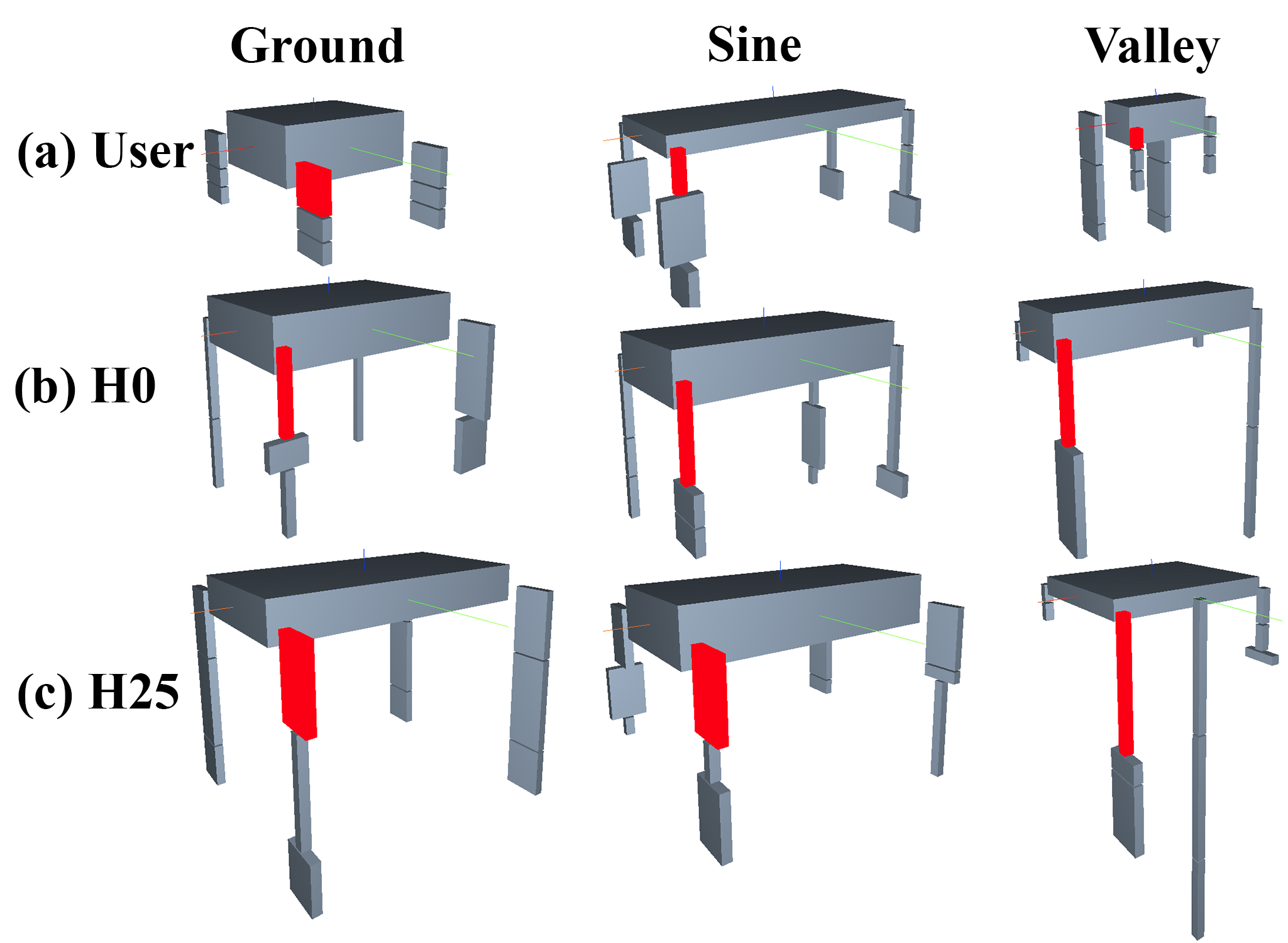}
    \caption{Example robots: (a) Human designs included in the initial
    population of H25. (b) High-performing designs in final population of H0.
    (c) High-performing designs in final population of H25. Red marks the left front
    leg.}
    \label{fig:example_of_generated_robots}
\end{figure}

Here, we investigate the impact of injecting human designs (and thus human
creativity, intuition, and expertise) into the initial population of an ER
experiment that produces legged robots
(\figurename~\ref{fig:example_of_generated_robots}).  In this way, the human
designer can influence the search for solutions that satisfy design
requirements, whilst also exploiting the design space coverage afforded by
evolutionary search.  We implement a quality-diversity \cite{pugh2016quality}
evolutionary algorithm based on MAP-Elites \cite{mouret2015illuminating},
providing an effective performance-design space map.  We couple this to the
Robogami \cite{doi:10.1177/0278364917723465} design tool, which enables
interactive human-led design exploration and provides a straightforward way to
integrate human designs into the evolutionary process.

We hypothesise that the inclusion of human designs in an ER experiment affects
the running progress and the final result of the algorithm, either in terms of
the performance of the evolved robots, the coverage of the map, or both.

MAP-Elites is typically initialized with a purely random sample of the design
space, which makes the initial population highly diverse across the feature
space. This increases the likelihood that the high-performing region would be
explored and illuminated. We observe, however, that for many design tasks, a
human designer will already have some intuition for what a successful design
looks like.  They may also approach a design problem from creative directions.
Both of these features can be used to enhance a design process by seeding the
initial population with more focus on the high-performing regions, especially
when the regions are at corners.  The better the human designer's intuition is,
the closer the initial population will be to the optimal region, and the less
evaluations it will take to illuminate the region and surrounding areas.  At the
same time, if the design task is complex or difficult for human designers, their
intuition may actually be detrimental to the search since the designs may not
only fail to bring the initial population closer to the optimal region, but
sacrifice diversity as well. We seek to quantify these effects.

We focus the study mainly on two research questions: \textbf{RQ1} What effects
would human input have on the intermediate population when algorithm is running?
\textbf{RQ2} What effects would the human input have on the final population?

First, we collect human designs across a range of terrains using a visual design
interface (\figurename~\ref{fig:robot_design_ui}).  Humans are tasked to design
robot morphologies that generate high-performance locomotion behaviour across a
range of increasingly difficult (for the robot) and decreasingly intuitive (for
the designer) terrains (\figurename~\ref{fig:environments}).  In stage 2, we
conduct a range of experiments that use varying ratios of human and
randomly-initialised designs. We assess the effect of changing ratios of human
and random designs in the initial population.  Overall, results indicate that
human inputs could have either positive or negative impact on the final evolved
robot fitness, depending on the quality of the designs, and a negative impact on
early stage map coverage. Optimal ratios of human and random designs are shown
to depend on environmental complexity.

The main contributions of this work are (i) the first work that uses human
designs to evolve robot morphologies, (ii) extensions to the Robogami software
that allow for coupling to MAP-Elites and a multi-participant user study, and
(iii) a detailed analysis of the effects of adding human inputs to the initial
population.

\section{background}
\label{sec:background}
\subsection{Computational Co-design}
Human-influenced computational design aims to combine compute-based rapid
assessment with human intuition and expert knowledge to collaboratively explore
and optimise within a given design space.  Typical goals include reducing
designer effort and reaching parts of the design space that are otherwise
difficult to access.  Applications are diverse, including optimising CAD models
\cite{schulz2017interactive}, designing model aeroplanes
\cite{umetani2014pteromys}, and assisting in the sketching of levels for
computer games \cite{liapis2013sentient}.  Visualisation plays an important
role, particularly when dealing with indirect solution representations which may
be unintuitive to explore without visual and performance feedback
\cite{hoisl2012visual}.  Works closely related to ER include interactive design
of 3D printable mechanical characters \cite{10.1145/2461912.2461953} that
reproduce desired animatronic motions through the computationally-assisted
placement of actuation mechanisms, and robotic creatures
\cite{10.1145/2816795.2818137} including legged robots
\cite{desai2018interactive}, where computational design is used to generate a
plausible gait for a given hand-designed morphology.  In all cases, the human
predominantly leads and controls the design process, with the computational
element used in a supporting role, e.g., to generate suggestions, verify the
design, or support the designer by creating working actuation for their designs.

\subsection{Interactive Evolutionary Algorithms}
Human input can be integrated into an Evolutionary Algorithm (EA) in several
ways.  For example, expert knowledge can be used to set parameter limits and to
design fitness functions~\cite{howard2020diversity}. Typically, the only
feedback the user receives on their choices is at the end of the experiment,
when solutions can be analysed to see if these settings produced the intended
results.  Interactive Evolution \cite{banzhaf2000interactive} is concerned with
more in-depth interplay between user input and evolutionary processes.  Early
examples include Sims' `Galapagos' exhibit where a human's interest (measured by
the time spent looking at a specific screen displaying an evolved art piece) was
used to drive the evolutionary process, and indirectly-represented digital art
\cite{mccormack1993interactive}, as well as the interactive evolution of
dynamical systems \cite{sims1992interactive}.  Interactive EAs have applications
as diverse as molecular design \cite{lameijer2006molecule} to evolution of
digital images \cite{secretan2008picbreeder} and brochures
\cite{c701f700929e47fb9e7b706bd149000e} to game level design
\cite{cardamone2011interactive,10.1145/3377930.3389821}.  Work on exploring the
design space around a provided CAD model \cite{UploadAnyObject}, e.g., injecting
user designs and then evolving them, demonstrates the benefits of mixing user
inputs and evolution.
We also note the success of interactive evolution in the domain of parametric
design \cite{doi:10.1177/1478077118778579}, which is our target domain.

\subsubsection{Interactive Quality-Diversity}
Quality-Diversity (QD) algorithms \cite{pugh2016quality} are a family of
evolutionary algorithms that aim to produce a wide range of high-quality
designs, with two popular variants being NSLC \cite{lehman2011evolving} (using
Pareto optimisation to maintain diversity) and MAP-Elites
\cite{mouret2015illuminating} (which uses a discretised feature map). QD is
particularly suited to design tasks, as a wide variety of performant designs are
encouraged to effectively map out an entire design space \cite{gaier2018data},
generating valuable feedback to the designer and identifying regions where
interesting solutions might lie. MAP-Elites has also been investigated in the
context of interactive evolution, where game level designs can be selected from
the map and edited by the user during an evolutionary run
\cite{Alvarez_Font_Dahlskog_Togelius_2021}, with preliminary work showing the
ability for human designs to unlock previously undiscovered areas of the design
space.

\subsubsection{Interactive Evolutionary Robotics}
Evolutionary Robotics (ER) (see numerous overviews e.g.,
\cite{bongard2013evolutionary,doncieux2015evolutionary,nolfi2000evolutionary})
studies the automatic generation of robot morphology and control within a given
environment, primarily to generate suitable behaviours, or as a tool to study
theories of embodiment, e.g., \cite{howard2019evolving}.  Originally simulated
\cite{sims1994evolving}, ER is now frequently associated with fabrication and
physical instantiation of both rigid \cite{lipson2000automatic} and soft robots
\cite{howard2021getting} via 3D printing.  Robogen is a pertinent UI/simulation
enabled evolutionary robotics platform~\cite{auerbach2014robogen}.  In our work,
we use the Robogami software, which was designed for user interaction and which
offers a direct route to eventual fabrication-based studies through 3D printing
or folding of the resulting robots through autogenerated fabrication plans.
\emph{Interactive} evolutionary robotics is sparsely covered in the literature.
ER automation of certain design tasks can reduce barriers to robot design
\cite{10.1145/2330784.2330955}, for example, by handling controller optimisation
of non-adaptive morphology, by using controller input to add new rules
on-the-fly for an evolved classifier-based robot controller \cite{1043879}, and
interactive approaches based on cellular representations \cite{ed1996cellular}.
To the best of our knowledge, interactive evolutionary morphology generation
does not appear in the literature.

\subsection{Literature Summary and Motivation}
Our work sits at the intersection of Evolutionary Robotics and Computational
Co-Design: users interactively design solutions, which are then harnessed to
improve an evolutionary process.  Rather than tweaking computed designs, users
iteratively tweak their own designs based on fitness and visual behavioural
feedback from the simulator.  This paper details the first such experiment in an
evolutionary robotics context.   Our approach combines the benefits of evolved
and human designs in a way that does not require the user to constantly interact
with the design software to 'tweak' designs, nor stop the evolutionary process
to wait for user input.  Compared to conventional computational co-design, the
emphasis for discovery is shifted more onto the computational (evolutionary)
element, which is responsible for ultimately evolving high-performance solutions
yet is guided by its given input designs.

\section{System Overview}
\label{sec:system_overview}
\begin{table}[tb]
    \centering
    \small
    \caption{MAP-Elites Parameters (heuristically determined)}
    \begin{tabular}{c|c}
        \textbf{Name} & \textbf{Value} \\
        \hline
         initial population size & 30 \\
         number of iterations & 2000 \\
         evaluations per iteration (batch size) & 30 \\
         archive map size & 20 $\times$ 20 \\
         crossover rate & 0.75 \\
         mutation rate & 0.1
    \end{tabular}
    \label{tab:map_elites_param}
\end{table}
Our study builds a closed-loop system that let users first build robots via a
interactive design tool backed by the Robogami
software~\cite{doi:10.1177/0278364917723465}, then use a
MAP-Elites~\cite{mouret2015illuminating} based searcher to evolve their designs
and generate better solutions.

\subsection{Robogami Designer}
\label{ssec:robogami_designer}
Robogami is a interactive robotic design tool based on a design-by-composition
framework that let users design robots by composing independently manufacturable
robot parts together. The software also includes algorithm that provides
interactive feedback to users, guiding their exploration by checking validity
and manufacturability of the design at each step. This provides a novice user a
smooth approach to get started and generate a design that reflects their wish.
To integrate this front-end to our system loop, we (1) simplify the user
interface and add design constraints to the software to limit the design space
as described in Section~\ref{ssec:robot_representation}; (2) link the software
to a simulator described in Section~\ref{ssec:task_and_robot_fitness} that
evaluates the design's fitness and provides user with visual feedback; (3)
develop a compiler that compiles the user design to the design vector that can
be loaded by the MAP-Elites searcher described in
Section~\ref{ssec:map-elites_seacher}; (4) wrap the software with a user study
guiding system that helps the users go through the user study described in
Section~\ref{sec:user_study} with minimum interruption by the researchers.

\begin{figure*}[tb]
    \centering
    \includegraphics[width=\linewidth]{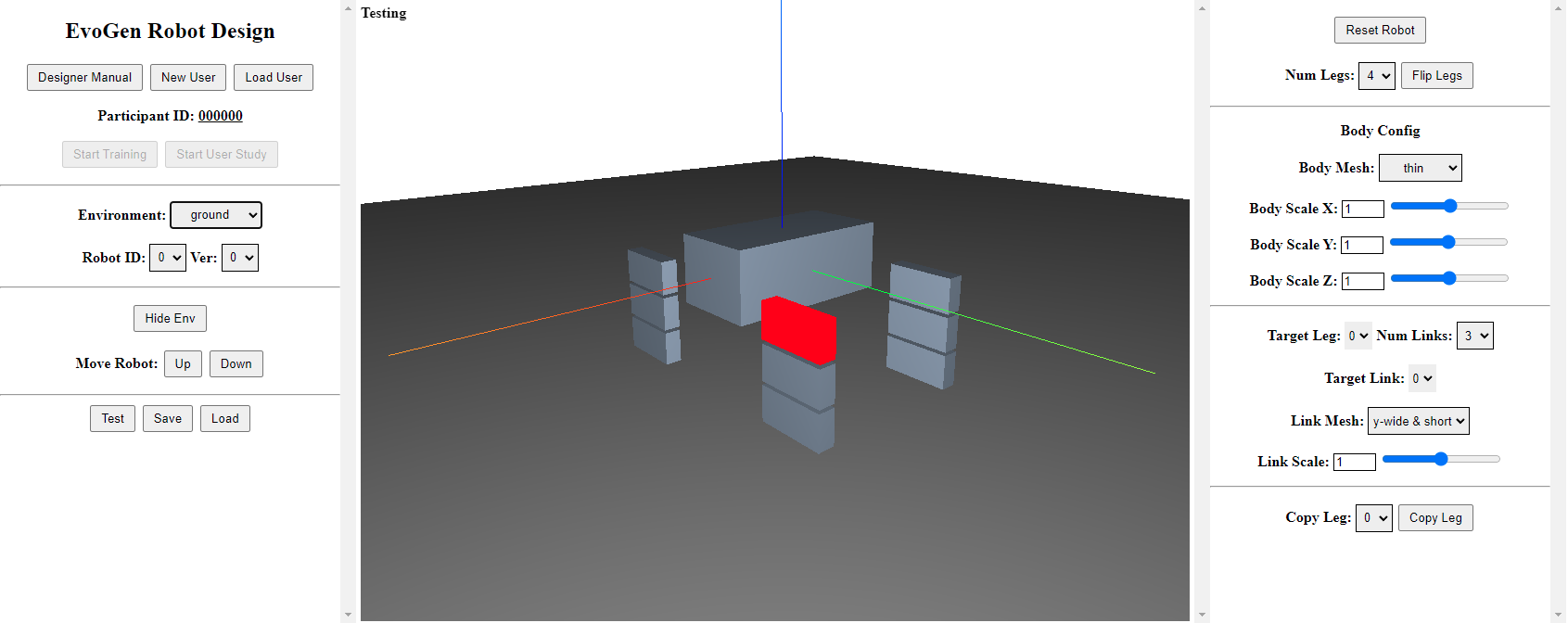}
    \caption{GUI of the robot designer. User  manipulates parameters in the
    genome vector via options on the right. The resulting design is displayed
    in the center. Options on the left  change the environment, load a pre-existing
    design, and simulate the current design. When triggered, an animation of the
    robot's walking behavior and the final fitness value are shown to the user.}
    \label{fig:robot_design_ui}
\end{figure*}

The resulted UI is shown in \figurename~\ref{fig:robot_design_ui}.

\subsection{Robot Morphology and Gaits}
\label{ssec:robot_representation}
ER uses a range of robot representations, which can be classified as direct or
indirect.  The choice of genome vector dictates the types and variety of robot
designs that can be achieved. More descriptive genomes permit a larger design
space, potentially improving the quality of the output design. However, larger
genomes increase training time as a larger design space must be explored.  Our
choice must balance genome complexity and design space expressiveness with the
ability for changes in genome parameters to be easily understood by
corresponding changes in the robot to aid in human design exploration. As such,
we use a direct representation that captures these features.

\begin{figure*}[t]
    \centering
    \begin{minipage}{0.65\linewidth}
        \subfloat[]{\label{fig:design_vector}\includegraphics[width=\linewidth]{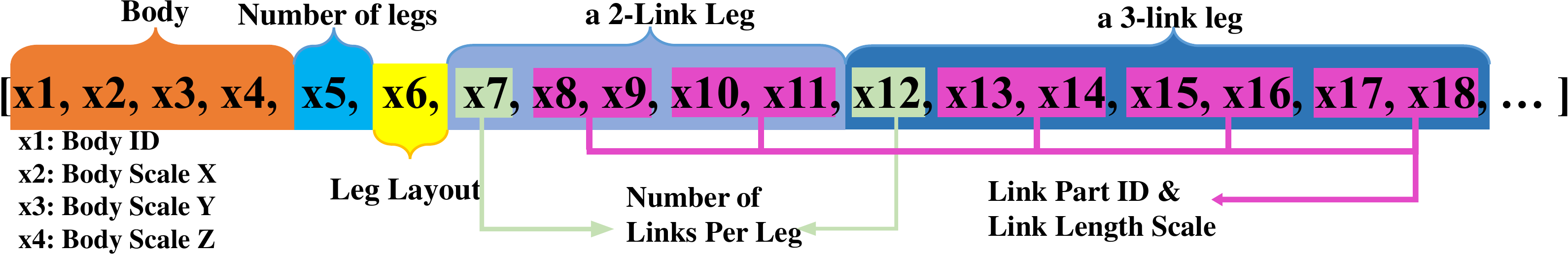}} \\
        \subfloat[]{\small
        \begin{tabular}{c|c}
            \textbf{Name} & \textbf{Range} \\
            \hline
            Body Shape ID & Integer 1 - 6 \\
            Body Scale $x$, $y$, $z$ & Double [0.5, 1.5] \\
            Number of Legs & Integer 2 - 6 \\
            Leg Layout & Bool {true, false} \\
            Number of Links & Integer 2, 3 \\
            Link Shape ID & Integer 1 - 7 \\
            Link Length Scale & Double [0.5, 1.5]
        \end{tabular}
        }
        \subfloat[]{\small
        \begin{tabular}{c|c c c}
            \textbf{ID} & \textbf{X} & \textbf{Y} & \textbf{Z} \\
            \hline
            1 & 10 & 10 & 4 \\
            2 & 15 & 10 & 4 \\
            3 & 20 & 10 & 4 \\
            4 & 10 & 5 & 4 \\
            5 & 15 & 5 & 4 \\
            6 & 7 & 5 & 4 \\
        \end{tabular}
        }
        \subfloat[]{\small
        \begin{tabular}{c|c c c}
            \textbf{ID} & \textbf{X} & \textbf{Y} & \textbf{Z} \\
            \hline
            1 & 1 & 1 & 4 \\
            2 & 4 & 1 & 4 \\
            3 & 1 & 4 & 4 \\
            4 & 1 & 4 & 2 \\
            5 & 1 & 4 & 7 \\
            6 & 1 & 1 & 7 \\
            7 & 1 & 1 & 10 \\
        \end{tabular}
        }
        \quad
        \subfloat[]{
            \begin{minipage}{0.1\linewidth}
            \includegraphics[width=\linewidth]{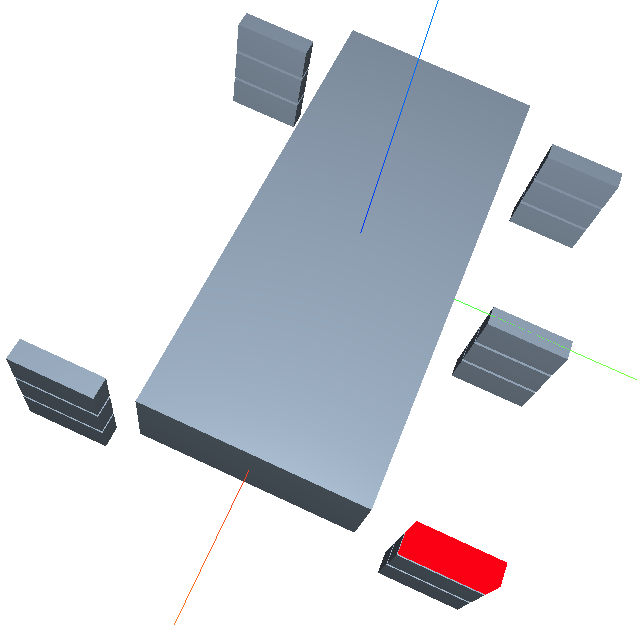}\\
            \includegraphics[width=\linewidth]{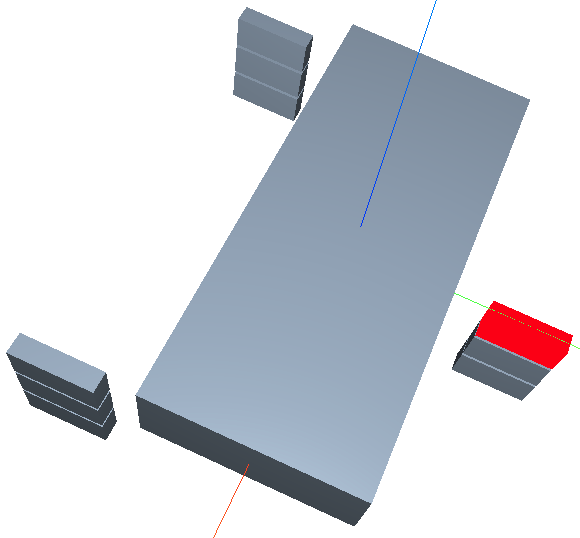}
            \end{minipage}
        }
    \end{minipage}
    \begin{minipage}{0.25\linewidth}
        \centering
        \subfloat[]{
            \label{fig:bodies_and_legs}
            \includegraphics[width=0.9\linewidth]{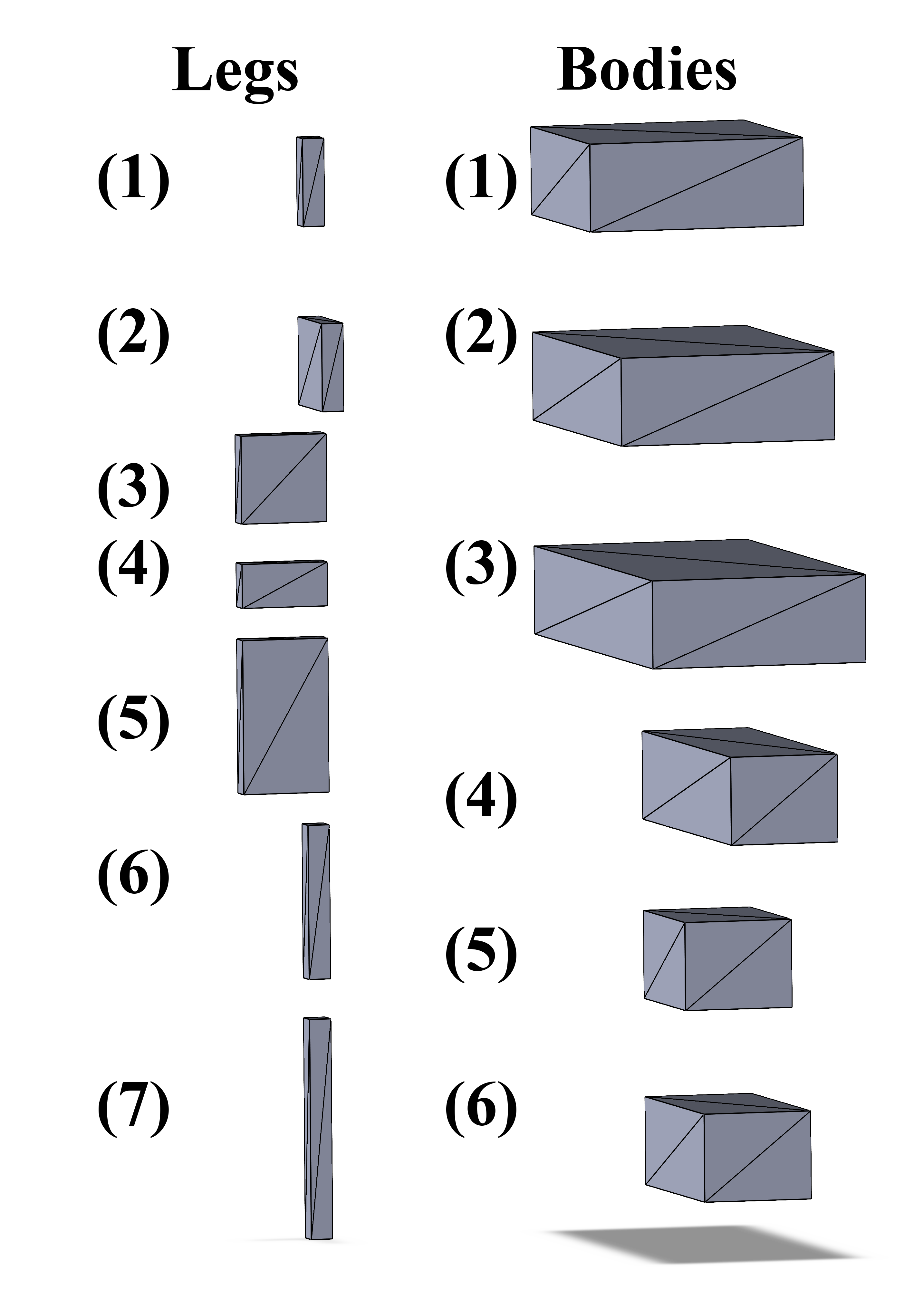}
        }
    \end{minipage}
    \caption{(a) Visual depiction of the robot genome, (b) allele definitions
    and ranges, (c) body part dimensions (cm), (d) body part dimensions (cm),
    (e) example robots highlighting the potential for asymmetric morphologies, (f)
    body and link shapes.}
    \label{fig:robot_representation}
\end{figure*}

In this study, we consider legged robots with 2-6 legs, each of which has 2 or 3
links. The genome, allele ranges, and mapping to a phenotype (robot) are
presented in \figurename~\ref{fig:design_vector}. The genome vector contains 17
to 53 variables depending on the complexity of the robot. Body Shape ID and Link
Shape ID correspond to rectangular prisms of different aspect ratios, and the
Body Scale and Link Scale are multipliers directly applied to the corresponding
dimensions of the parts, providing a flexible and scalable design space suitable
for both human and evolutionary exploration.

Legs are evenly distributed on each side of the robot's body and attached to the
middle point of the $z$ direction (See \figurename~\ref{fig:environments} for
direction definitions). A layout-mirroring flag controls which side of the robot
has more legs when an odd number of total legs is inputted. This flag is ignored
for robots with even number of legs. Leg links are connected to each other in
series with joints located at either end of the link. All joints are revolute
joints with axes of rotation parallel to the $y$ axis.  The robot parts are
assumed all to be of uniform density (2.5 g/cm$^3$), with dimension details
shown in \figurename~\ref{fig:robot_representation}.

\begin{figure}[t!]
    \centering
    \includegraphics[width=0.8\linewidth]{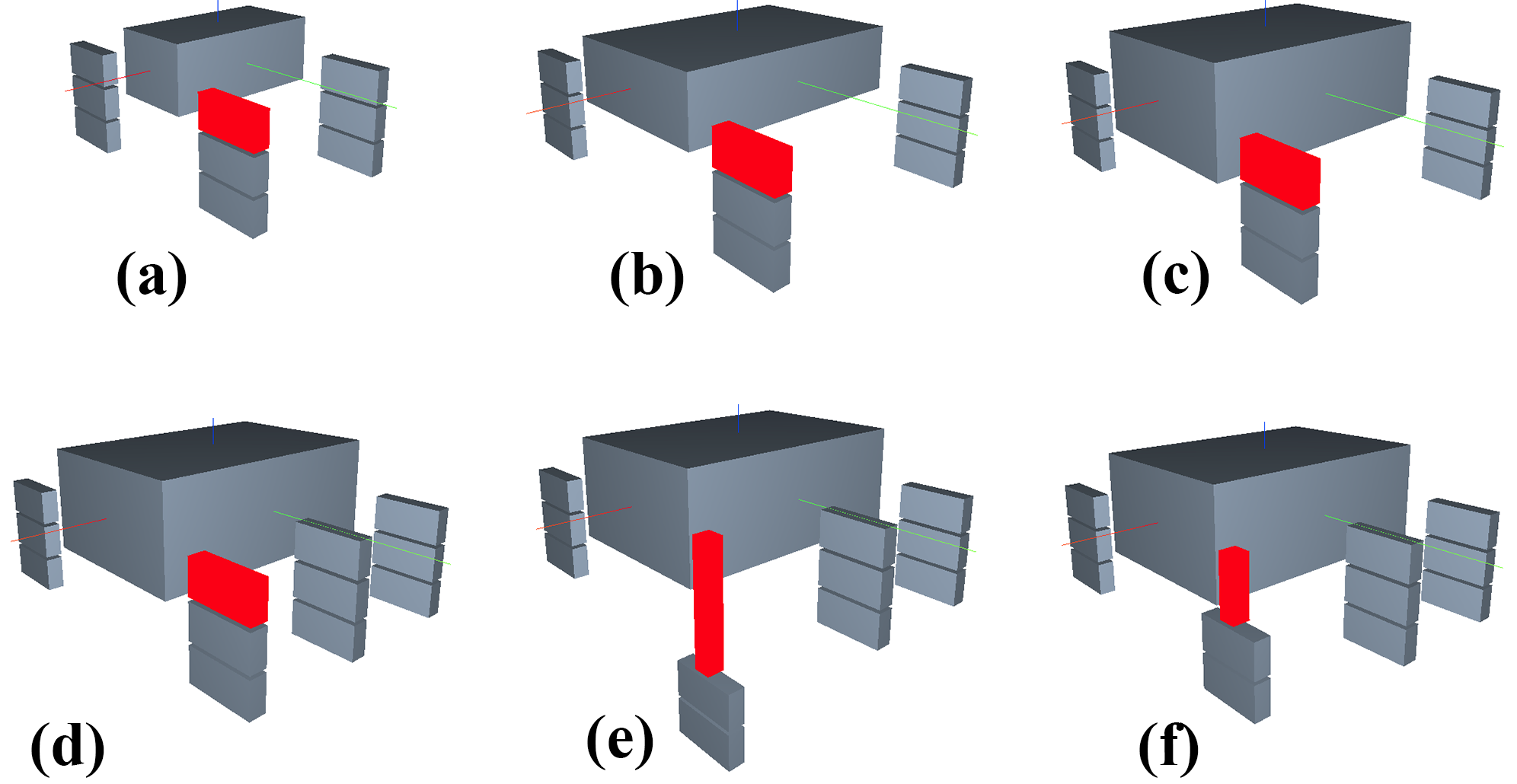}
    \caption{Example mutations, starting from (a) a 'neutral' quadruped design,
    that alters (b) body part, (c) body scale (1.5 times on z direction), (d)
    number of legs, (e) link part, (f) link length. Red marks the left front leg.}
    \label{fig:example_of_robots}
\end{figure}

Even with this simple robot representation, a variety of robot geometries and
kinematic structures can be achieved, including asymmetric robots with differing
numbers of legs or leg links on each side.
(\figurename~\ref{fig:example_of_robots}).

\begin{figure}[tb]
    \centering
    \includegraphics[width=\linewidth]{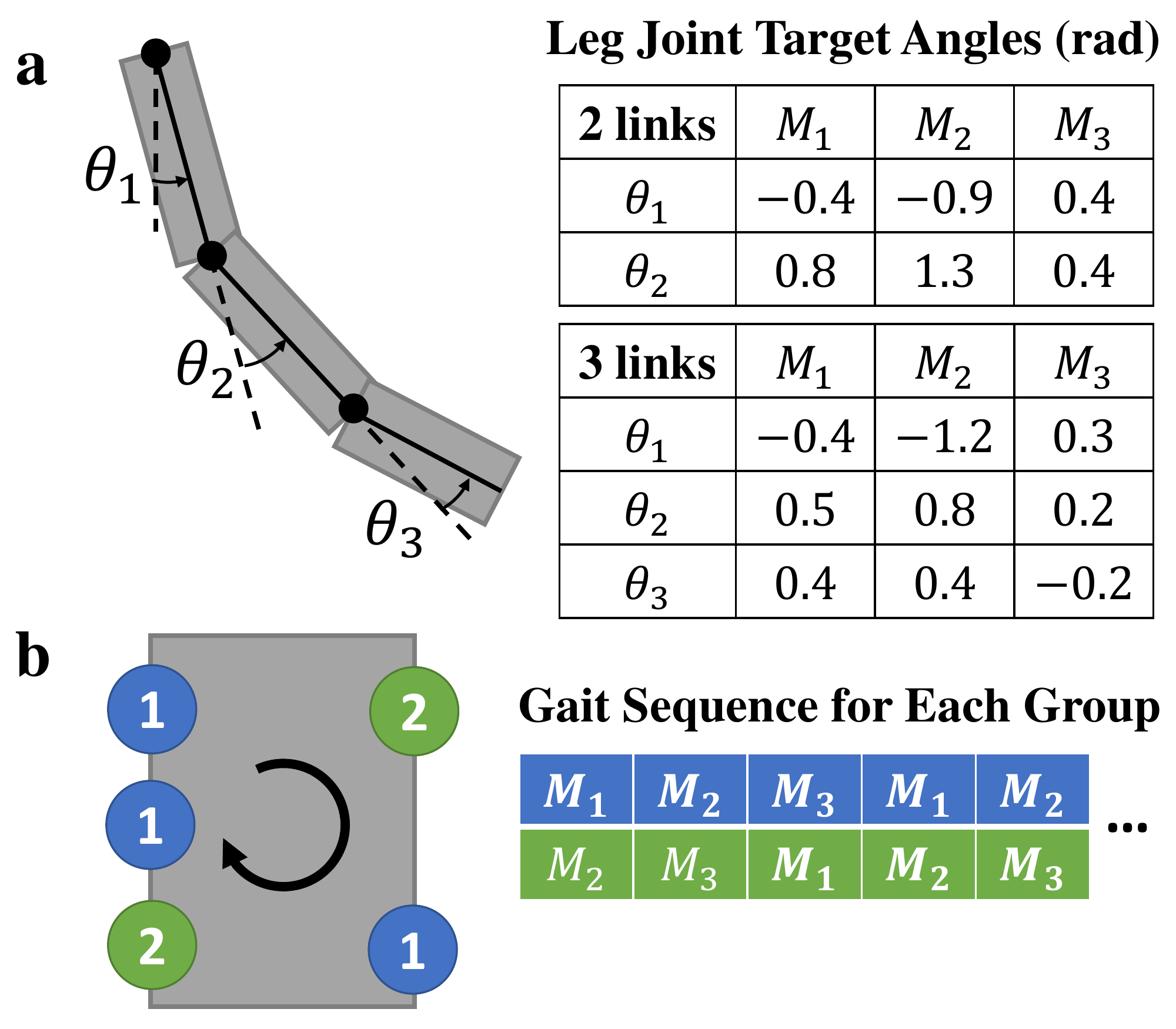}
    \caption{Gait executed by the robot designs. (a)~Target joint angles for
    legs of different numbers of links. (b)~Example grouping of legs and
    corresponding motion sequence. The front of the robot faces up.}
    \label{fig:gait}
\end{figure}

The gait of the robot is directly determined by the morphology, with a set of
predefined joint movements based on the design and layout of legs. Each of the
legs follows a motion sequence depending on the number of links in the leg
(\figurename~\ref{fig:gait}) with an offset depending on its placement on the
robot body. In particular, the legs are divided into two groups. Starting from
the front left leg, alternating group numbers are assigned in clockwise order.
Group 1 executes the motion sequence $M_1M_2M_3$ simultaneously as Group 2
executes $M_2M_3M_1$, and the process repeats over multiple gait cycles. The
group switches to the next joint target when all of the joint angles are within
0.01~rad of the target angle, or 3~s after the other group finishes its current
motion, whichever happens earlier. The joint angles and motion sequences
associated with the joints in the legs are kept constant over all robot designs;
that is, they are not included in the evolutionary design.  The gait is
controlled by a PI controller over both velocity ($k_p=10,k_i=0.3$) and position
($k_p=2,k_i=0$), with a maximum joint velocity of 2~rad/s.  A deterministic
controller was used to remove a potential level of indirection, in the form of
body-brain optimisation, from the user design problem, allowing a predictable
control output per morphology to simplify the design process.

\subsection{Task and Fitness}
\label{ssec:task_and_robot_fitness}
To make sure the participants of the user study could get good intuition on some
problems, we define the task for the robot to be simply walking forward in
different environments, and define the fitness to be heavily determined by the
distance traveled during the given time.

At the start of each simulation, the robot is placed at a fixed point at one end
of the environment, with $z$ position determined by having the foot of its
longest leg lifted above the terrain surface directly under it by 2~cm. The
robot is then simulated forward for 30~s of simulated time with simulation step
size of 0.005~s. We use Project Chrono~\cite{tasora2015chrono} as our simulator.

\begin{figure*}[tb]
    \centering
    {\includegraphics[width=0.32\linewidth]{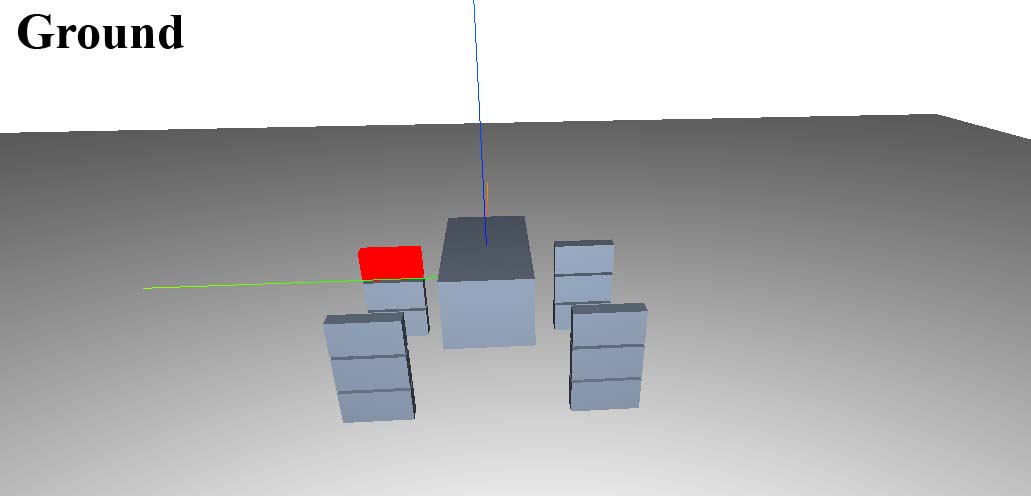}}\;
    {\includegraphics[width=0.32\linewidth]{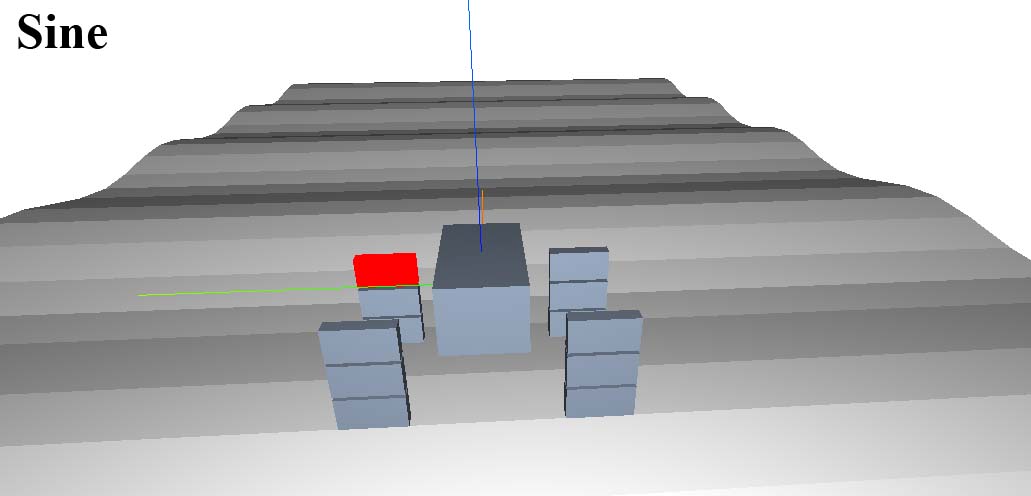}}\;
    {\includegraphics[width=0.32\linewidth]{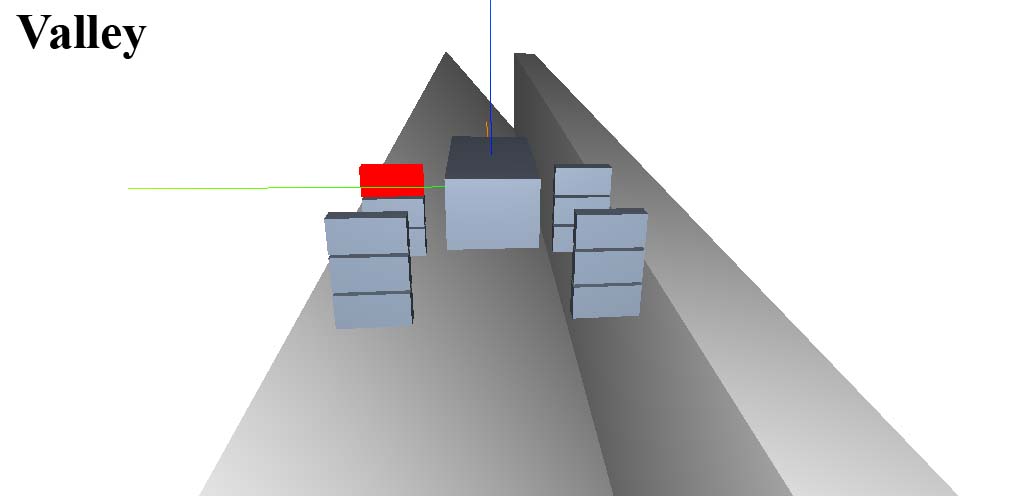}}
    \caption{Test environments: Ground (G), Sine (S), and asymmetric Valley (V)
    with one vertical wall and one wall at 45$^o$. Neutral robots provided for
    scale. Red marks the left front leg of robot. $+x$ points forward, $+y$ points
    leftward, $+z$ points upward}
    \label{fig:environments}
\end{figure*}

Since we expect the impact of human-created designs to differ depending on the
complexity of the design task and the quality of human intuition, we designed
three environments Ground (G), Sine (S), Valley (V)
(\figurename~\ref{fig:environments}) with increasing degrees of difficulty of
walking forward. Ground is expected to be the easiest task for a human as the
environment has little impact on robot behaviour.  Sine introduces additional
difficulty by requiring designers to think how navigate robots through the
bumps. Valley is the most difficult due to the introduced asymmetry, which is
counter to most human designers' intuition. The fitness objective $f$ is to
maximize the distance of the robot moving forward while minimizing divergence
from a straight line:
\begin{equation}
    \text{$f$} = \Delta x - 0.5 \ \text{abs}({\Delta y})
    \label{eq:robot_fitness}
\end{equation}
where $\Delta x$ is the net displacement in the forward direction and $\Delta y$
is the net displacement perpendicular to the forward direction at the time when
the simulation stops. If the system detects the robot has fallen off the edge of
the environment, the simulation terminates with fitness taken from the robot's
last position.

\subsection{MAP-Elites Searcher}
\label{ssec:map-elites_seacher}
We build the evolutionary searcher upon MAP-Elites, and link it to the simulator
described in Section~\ref{ssec:task_and_robot_fitness} for fitness and feature
evaluation. MAP-Elites holds its population in the archive map, which is a
discretized low-dimensional projective space of the search space. We use a 2D
map of $20 \times 20$ cells, with the 2 dimensions (features) being: (1)~the
length of the body along the $x$ direction (\figurename~\ref{fig:environments}
for direction definition), and (2)~the standard deviation of the leg lengths.
The length of the robot's body directly affects its separation of legs on the
forward direction, which then decides robot's maximum feasible leg length and
stability given our controller (\figurename~\ref{fig:gait}). The standard
deviation of the leg lengths quantifies asymmetry of the robot design.  These
two features were chosen through a small number of test runs using random
initial populations. Compared to other tested features, e.g., average length of
legs, maximum length of legs, total volume of robots, length of genome, etc.,
these two were found to be two of the most significant factors related to the
environments and fitness (Eq.~\ref{eq:robot_fitness}) we use.

We set the size of initial population to be 30 and run the algorithm for 2000
iterations. For each iteration, 30 parent designs are randomly selected from the
archive map to generate a batch of 30 children designs with mutation rate of 0.1
and crossover rate of 0.75. The children designs are evaluated according to a
fitness metric and projected back onto and stored in the archive map. If the
projected cell is occupied, or if two children designs are projected to the same
cell, the individual with the highest fitness is retained.

\section{Human Data Collection}
\label{sec:user_study}

\begin{figure}[tb]
    \centering
    \subfloat[]{
        \includegraphics[width=0.7\linewidth]{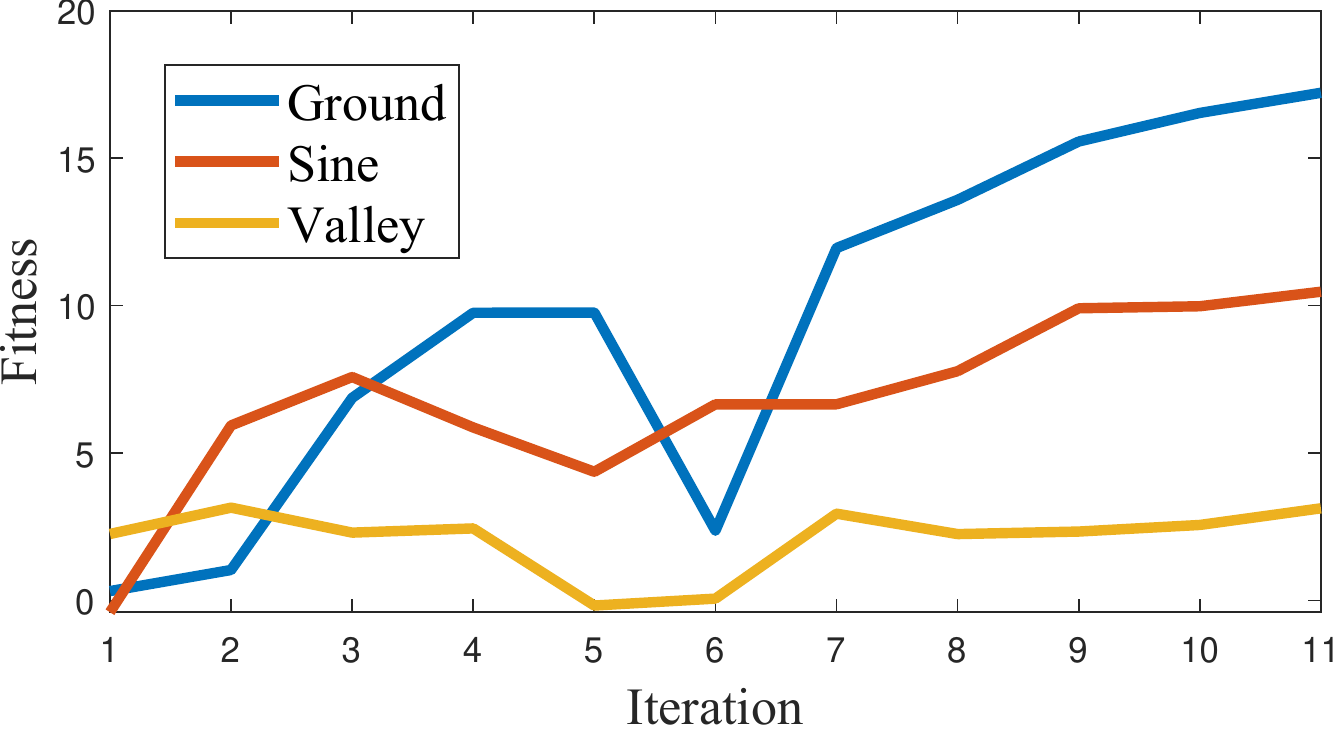}
        \label{fig:user_input_fitness}
    } \\
    \subfloat[]{
        \includegraphics[width=\linewidth]{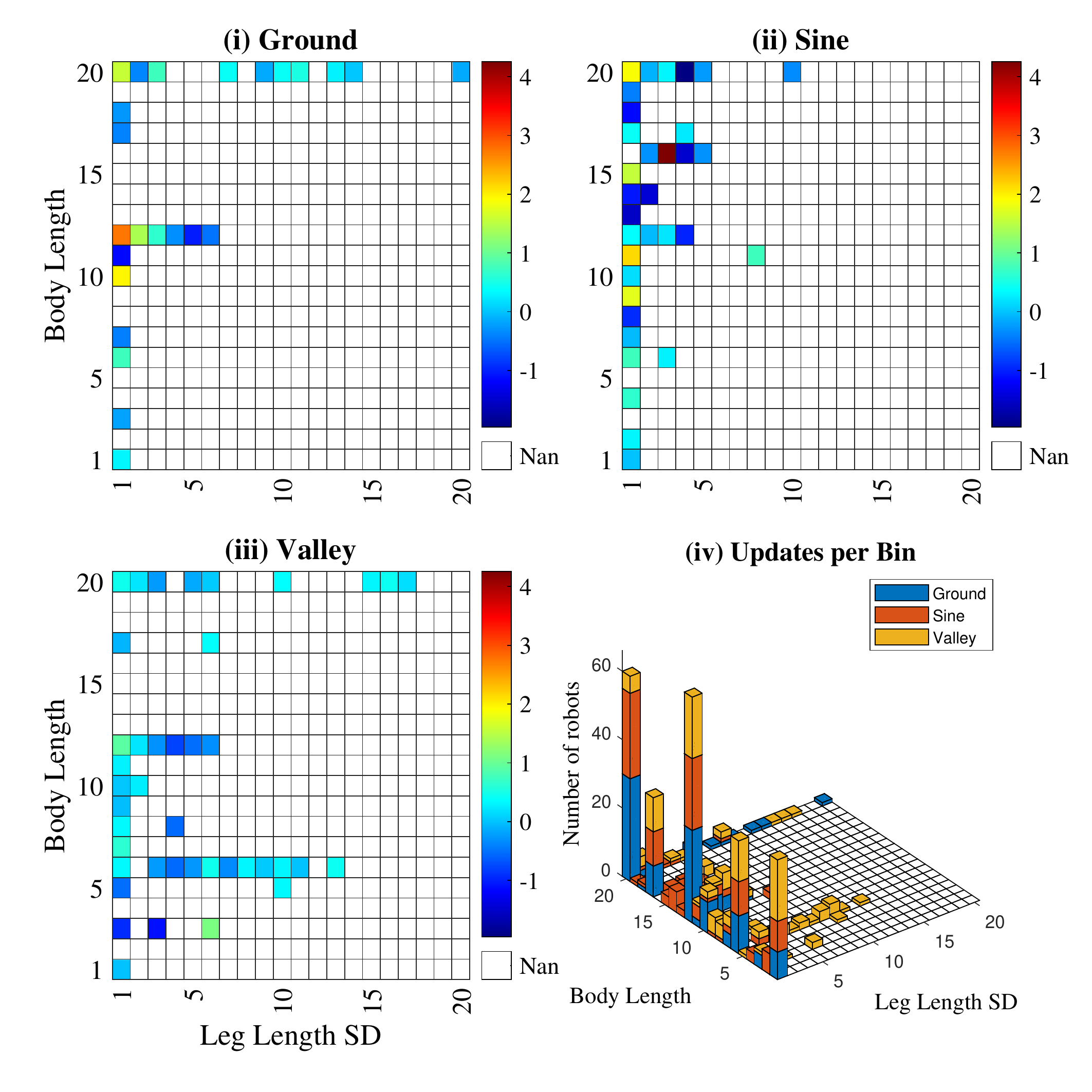}
        \label{fig:user_input_stats}
    }
    \caption{(a) Fitness progression over 10 user design iterations for one user
    on the three environments. (b) (i)-(iii) Archive map of all human designs
    per environment, coloured by fitness. (iv) Number of robots per bin on the map,
    coloured by environment type.}
\end{figure}

A user study\footnote{This user study has been approved by the IRB of University
of Pennsylvania under the title \textit{Robotic design decision making}
(protocol \#849907).} was conducted to collect human designs with the UI
described in Section~\ref{ssec:robogami_designer}. Data was collected from 13
students (5 undergraduate, 8 graduate) who all had some prior experience in
robot design. Each participant started with a 10-min. tutorial explanation of
the user interface on a sample design, then a 10-min. exploration session to
familiarize themselves with the system and ask questions, in which they could
design and test robots for an unlimited number of times on a different training
environment.  After the training, each participant was given the task and the
definition of fitness, and asked to design a robot for each of the three
environments.  At the start of each new environment, all options on the right
panel of the UI are set to the middle point of their range, showing the default
``neutral robot.'' The participant would adjust the parameters and design a
robot that reflects their intuition and knowledge, and send the robot to
simulation for walking visualisation and fitness.  Although the purpose of the
data collection was to gather designers' intuition about the optimal design, we
also wanted to remove the potential negative effects from the users'
unfamiliarity with the simulation and design software.  Each participant was
allowed to update the design 10~times per environment, and each update was
considered to include all of design changes made between two consecutive
simulation requests. Design genomes were saved at each iteration, resulting in
11 total designs per environment per user. Giving participants multiple chances
to test their design was to make sure they could fully express their intuition
and make the robot faithfully behave as they wish, without setting back by the
bias and barriers caused by the unfamiliarity to the system. And the
participants were not given unlimited trials either, so that they didn't
unintentionally evolve their designs by trial-and-error.

The order of the environments was randomized for each user to mitigate the bias
from one environment to the next, introduced by the increasing familiarity to
the system.

We expect the fitness of the design to improve over the iterations up until the
user is satisfied with their design quality, at which point they commit to their
design by simulating the same design for the remaining iterations or they test
minor tweaks.

A total of 143 designs (11 robots from 13 users) were collected for each
environment. In a post-processing step, we removed duplicates from when the user
requested a simulation multiple times for the same design, resulting in 125
total designs for the Ground, 128 for the Sine, and 130 for the Valley.

\figurename~\ref{fig:user_input_fitness}(b) shows the fitness progression of the
design iterations on different environments for one participant. As expected,
for the Ground and Sine, designs improved over the course of design iterations
for the first few iterations before fitness increases slow down near the final
value. The robot designs also undergo larger changes during earlier iterations
than later ones. The general fitness increases support the suitability of our
approach as a design tool.  In the Valley, however, the fitness achieved the by
the participant remained low and throughout the iterations and the designs
exhibited larger changes between iterations, indicating that users performed
more exploration for this task and that it was overall more difficult for them
to use their intuition on.

\figurename~\ref{fig:user_input_stats}(c) (i) - (iii) show the archive maps with
all designs from all users for the corresponding environment are inputted. A
large number of designs are concentrated on the left side of the map
corresponding to symmetric designs for all environments. Comparing to the sample
archive maps of the final population of H25 shown in
\figurename~\ref{fig:archive_maps}, we see the users successfully placed their
designs in the high-performing region for Ground and Sine, but not for the
Valley, indicating the less productive intuition for the latter.  And we see
more cells in \figurename~\ref{fig:user_input_stats}(c)(iii) are filled, as the
users were forced to explore more areas in the design space.
\figurename~\ref{fig:user_input_stats}(c)(iv) shows the number of robots that
goes into each cell in the feature map, with all robots included. The preference
of cells is another indicator that humans' exploration of the design space is
relatively condensed.

\section{Results}
\label{sec:results}
To explore \textbf{RQ1} and \textbf{RQ2}, we set up 5 test conditions with
varying numbers of human and random designs.
\begin{itemize}
    \item[(H0)] No human designs and 30 random robots.
    \item[(H5)] 5 human designs and 25 random robots. Human inputs were the top
         5 designs, with at most 1 design per user.
    \item[(H15)] 15 human designs and 15 random robots. Human inputs were the
         top 15 designs, with at most 2 designs per user.
    \item[(H25)] 25 human designs and 5 random robots. Human inputs were the top
         25 designs, with at most 3 designs per user.
    \item[(H30)] 30 human designs and 0 random robots. Human inputs were the top
         30 designs, with at most 3 designs per user.
\end{itemize}
Each test condition H$x$ is repeated 10 times with random designs uniformly
re-sampled for each repeat. The experiments are conducted on a desktop computer
with a quad core 10th Gen Intel i7 @ 3.4GHz CPU, taking approximately 6 hours
per run.

\begin{figure*}
    \centering
    \includegraphics[width=0.9\linewidth]{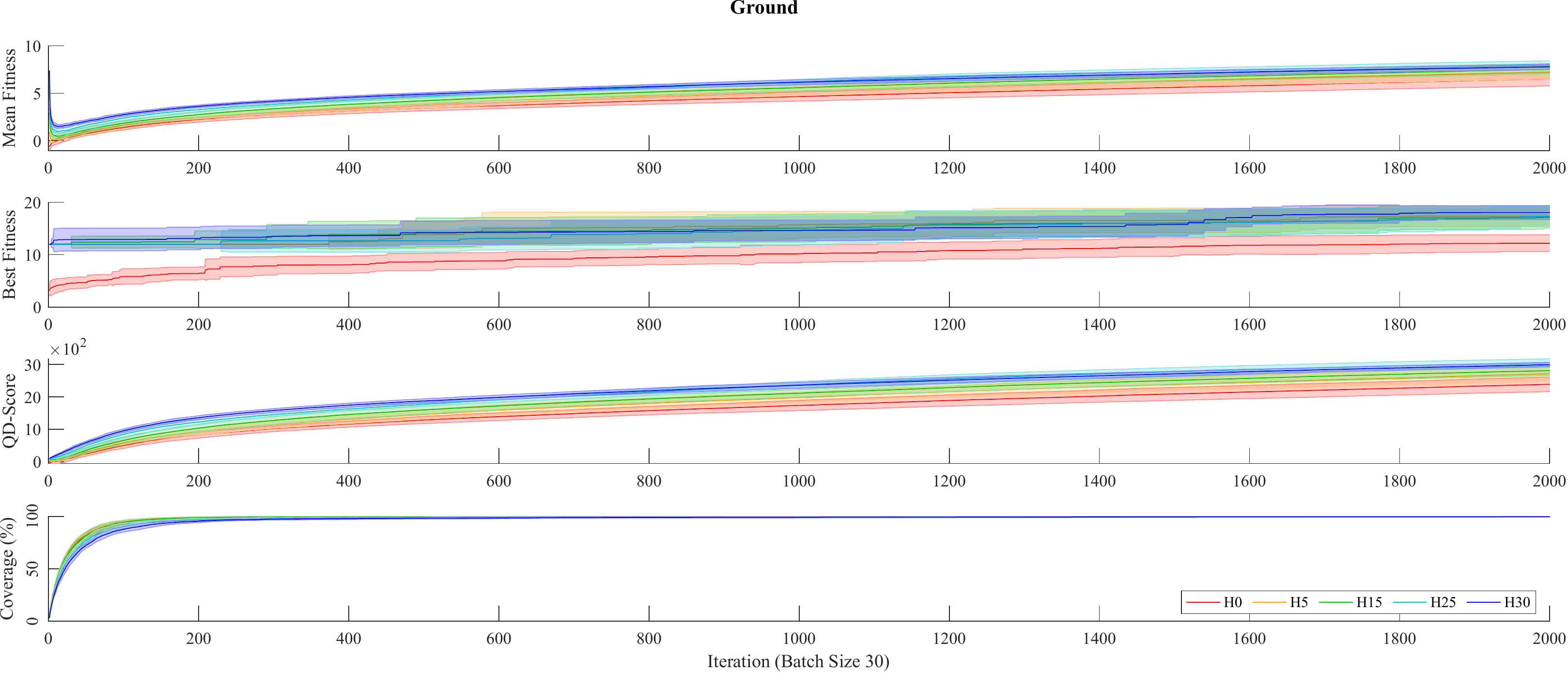}
    \includegraphics[width=0.9\linewidth]{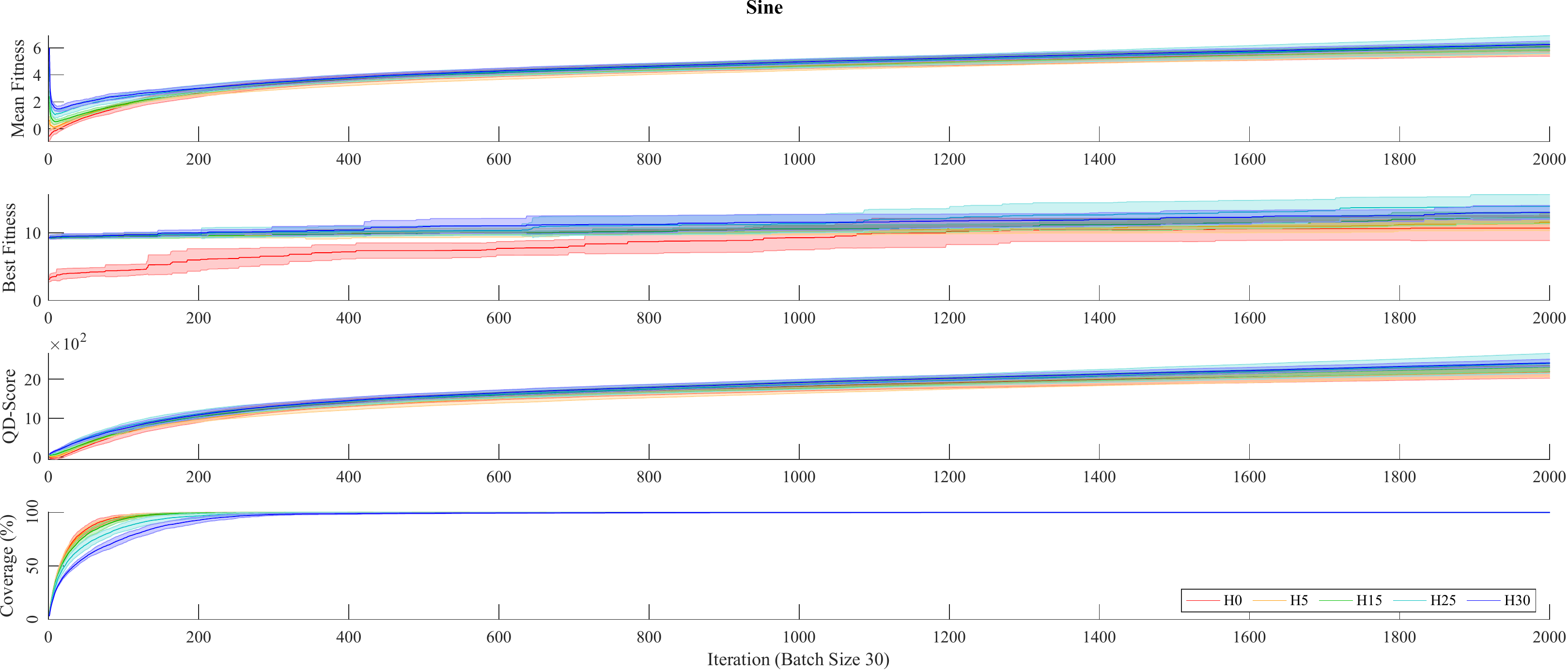}
    \includegraphics[width=0.9\linewidth]{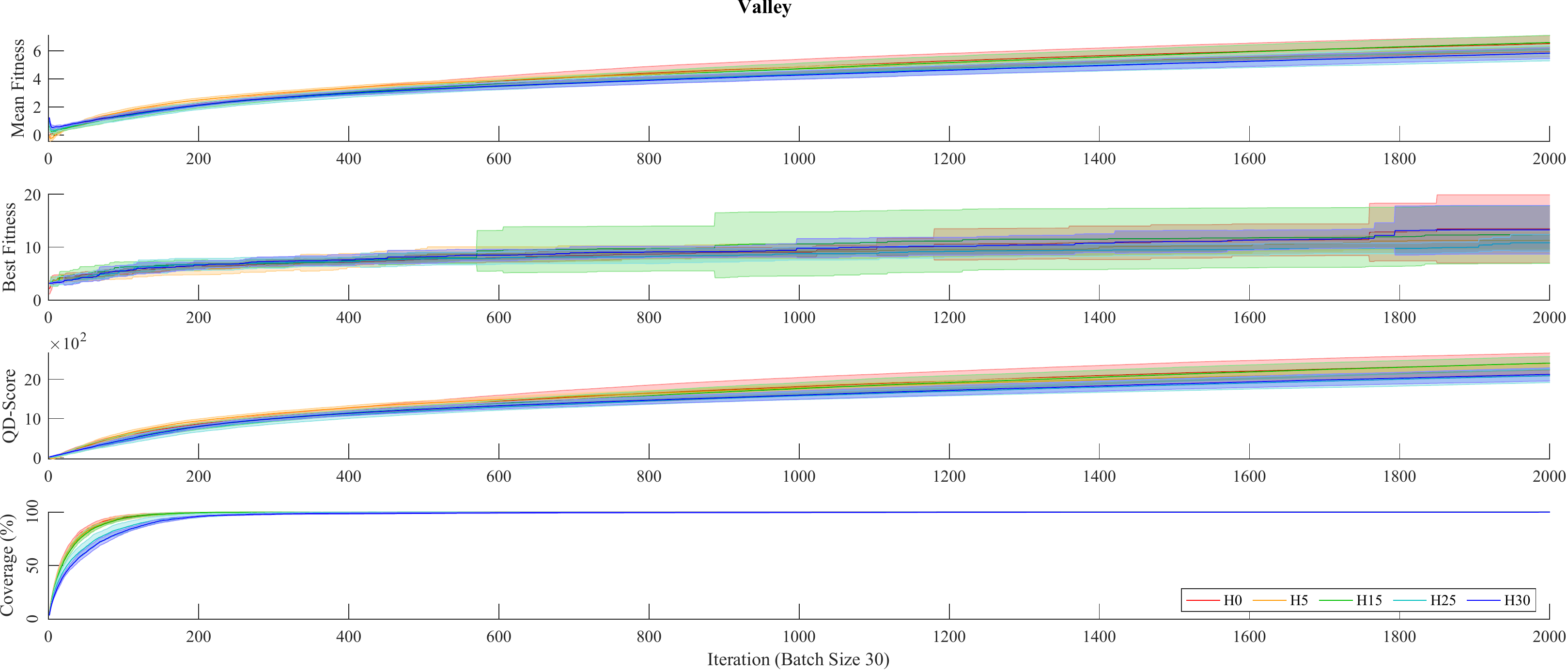} \caption{Plots
    showing statistics of intermediate population, including average archive
fitness, best single fitness of archive, QD-score coverage, as training
proceeds.}
    \label{fig:result_stat_plot}
\end{figure*}

\begin{figure*}[tb]
    \centering \subfloat[]{
        \label{fig:result_box_plot}
        \includegraphics[width=.3\linewidth]{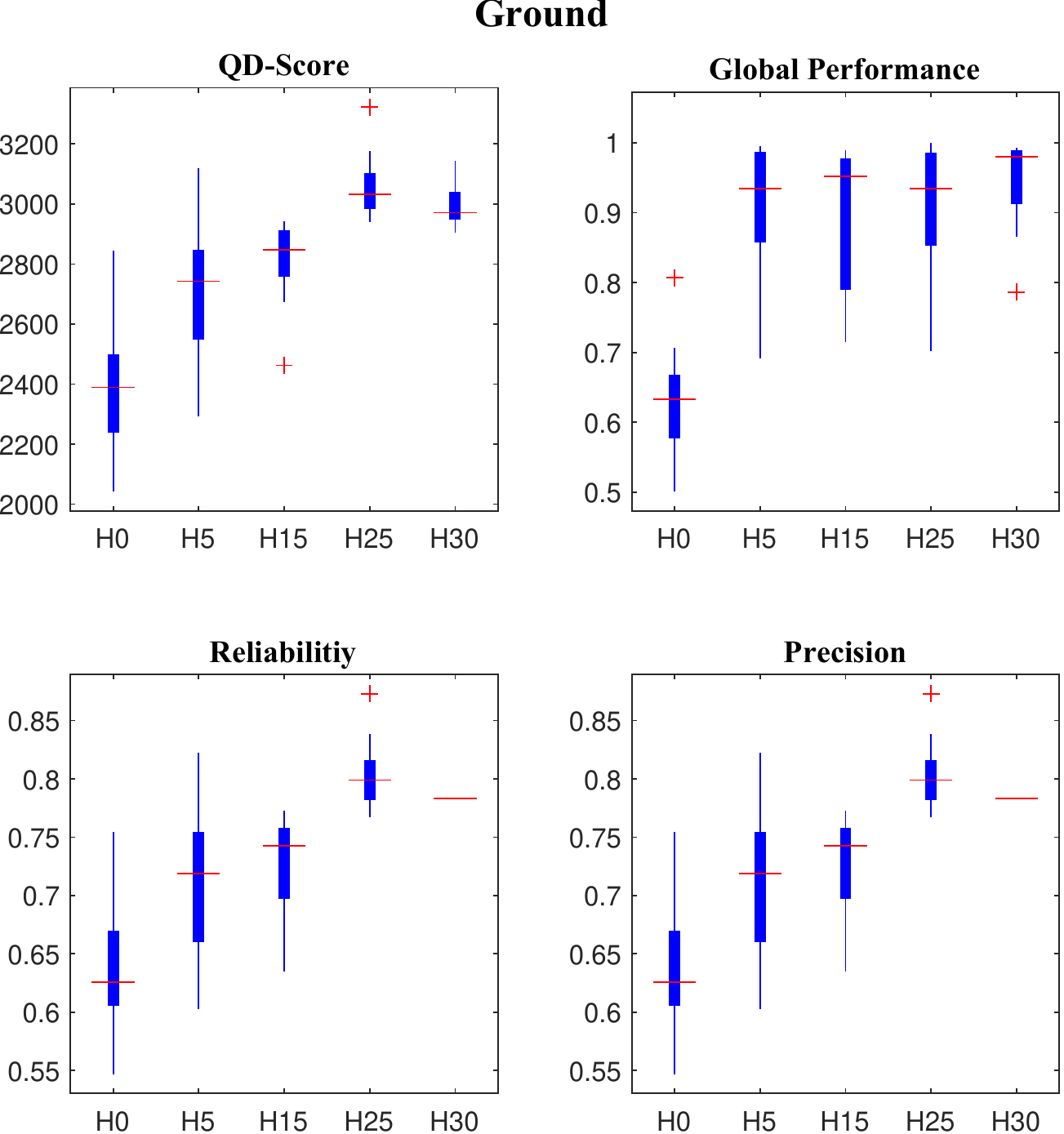} \quad
    \includegraphics[width=.3\linewidth]{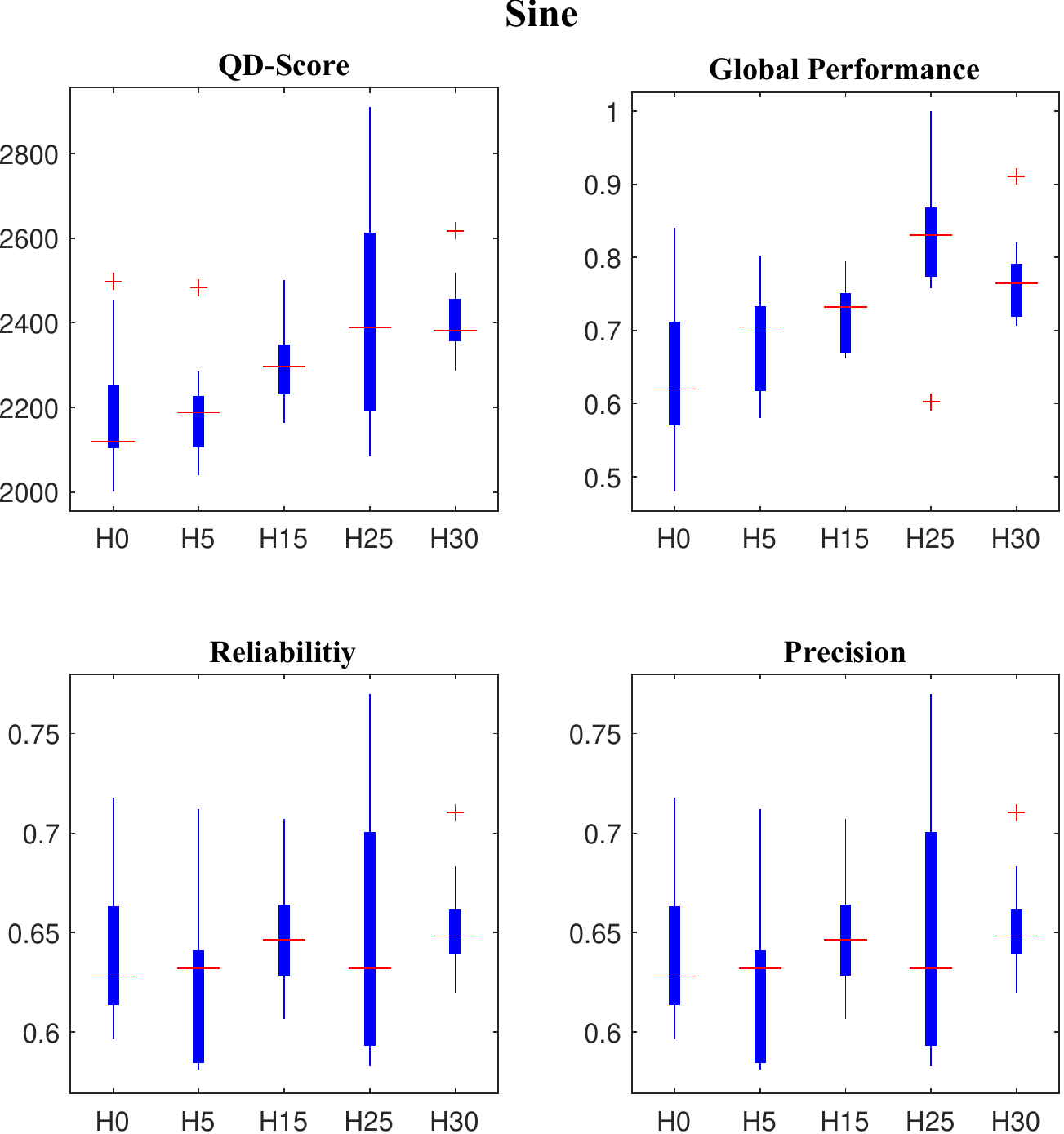} \quad
\includegraphics[width=.3\linewidth]{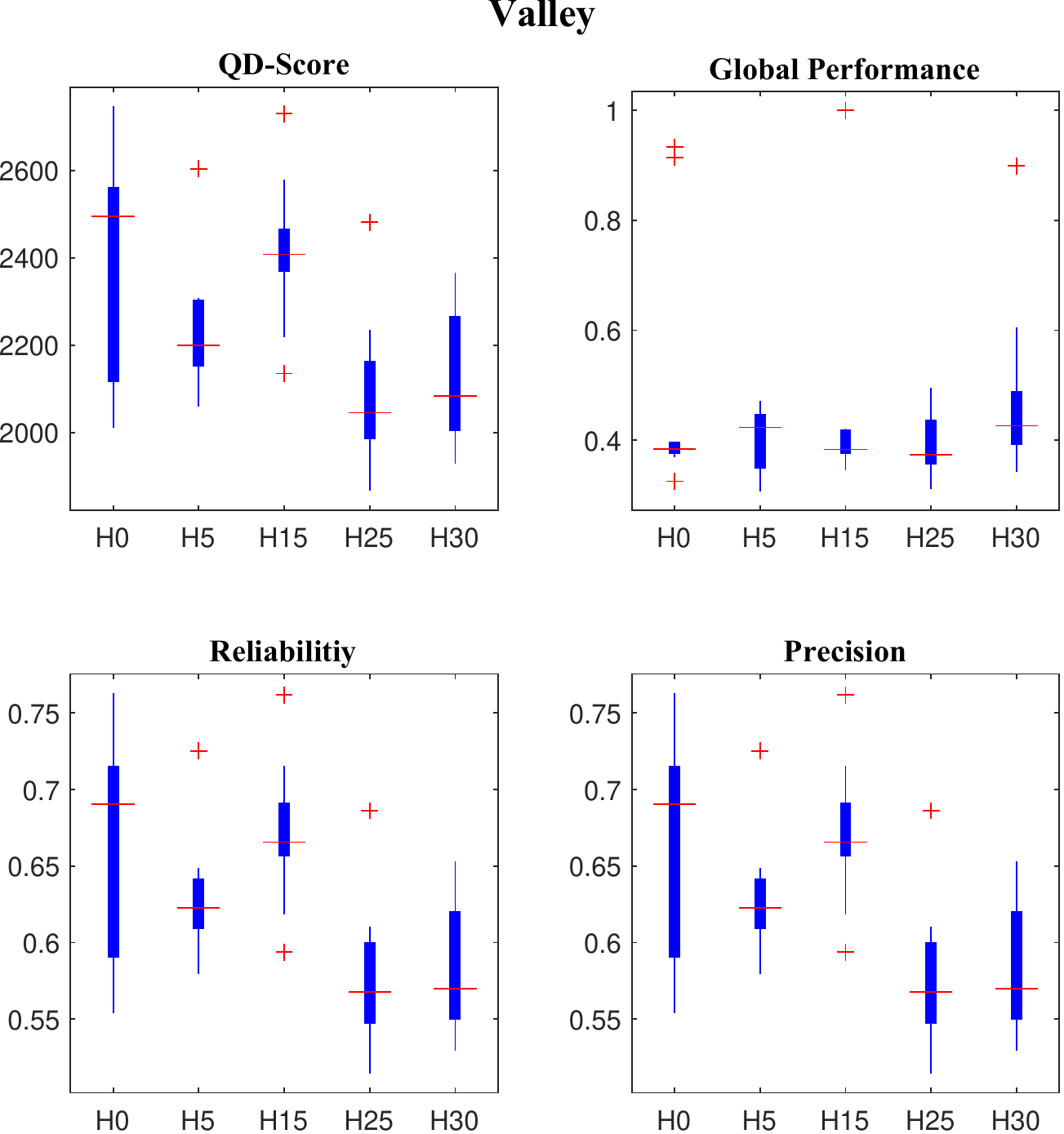} }
    \caption{Box plots showing statistics of final population, including
    QD-score, Global Performance, Reliability, Precision of the final archive of
each run. Red line marks median, ``$+$'' indicates outliers. Note the
reliability and precision plots are identical, as our experiment reaches 100\%
coverage in all runs.}
    \label{fig:result_figure}
\end{figure*}

\begin{figure*}[tb]
    \centering
    \subfloat[Ground]{\label{fig:archive_map_g}\includegraphics[width=0.3\linewidth]{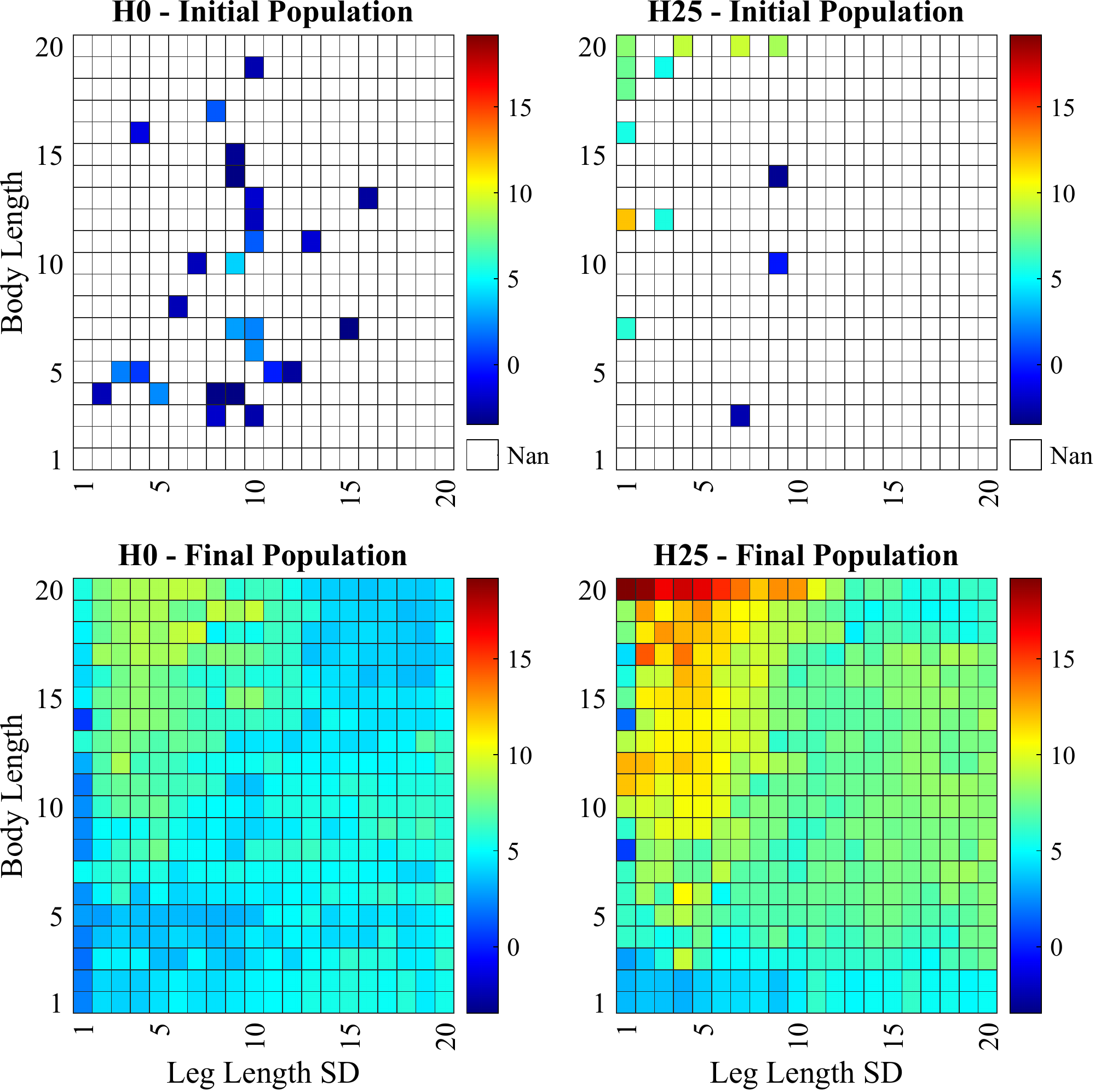}} 
    \qquad
    \subfloat[Sine]{\label{fig:archive_map_s}\includegraphics[width=0.3\linewidth]{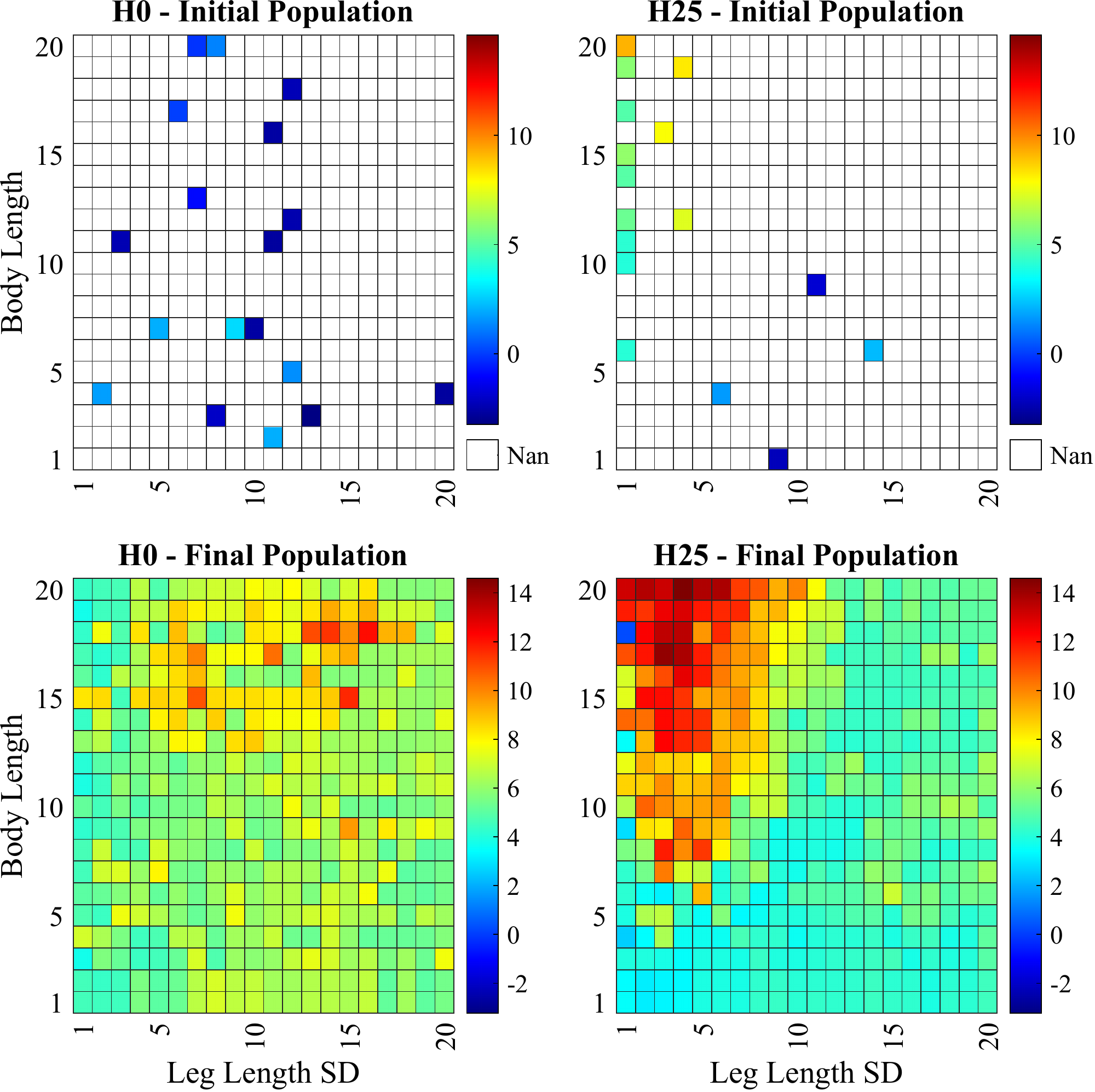}} 
    \qquad
    \subfloat[Valley]{\label{fig:archive_map_v}\includegraphics[width=0.3\linewidth]{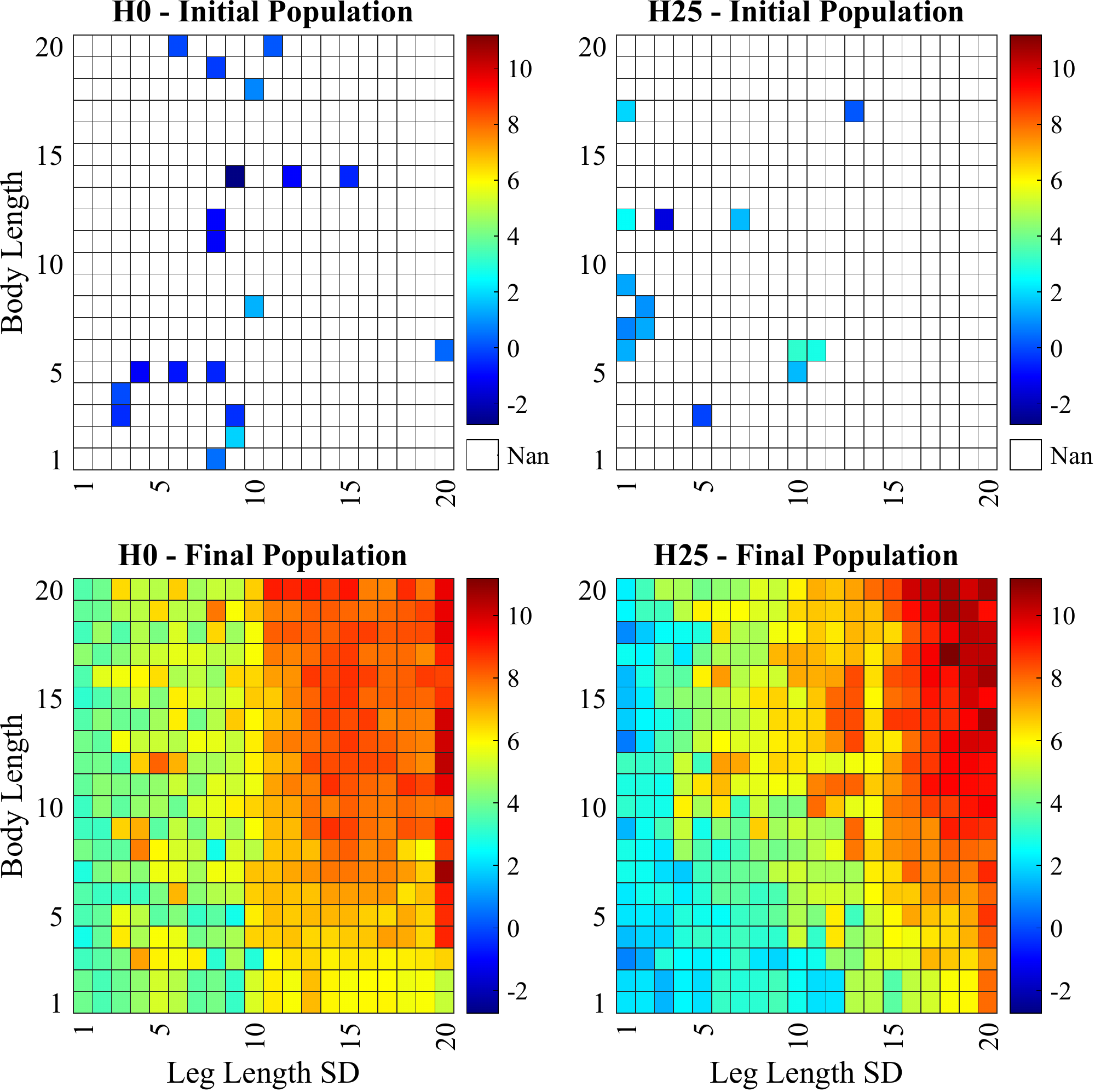}} 
    \caption{Representative maps resulting from a single run with a random
    initial population (H0) versus an initial population with 25 human designs
    (H25).  Colour represents fitness.}
    \label{fig:archive_maps}
\end{figure*}

The curves in \figurename~\ref{fig:result_stat_plot} show the transitional
changes of key metrics during the running of algorithm. Box plots in
\figurename~\ref{fig:result_box_plot} show key MAP-Elites metrics of the final
population.~\cite{mouret2015illuminating} \figurename~\ref{fig:archive_maps}
shows sample archive maps from H0 and H25 for three environments, and
\figurename~\ref{fig:example_of_generated_robots} shows initial and
high-performing robots sampled from those archive maps.
\figurename~\ref{tab:fitness_stat_test} shows the Mann-Whitney U-test result for
mean and best fitness of the final population of different test
conditions.\footnote{Supplementary data and resources are available at:\\
https://sung.seas.upenn.edu/publications/evorobogami-gecco-2022.}.

\begin{table}[tb]
    \caption{Mean fitness for the human designs, Initial (I) and final (F)
    population of H0 and H25. In the first column, the letters indicate the
    environment: G(round), S(ine), or V(alley). We also show the best fitness of
    user design, and the mean of best single fitness of different test runs (B).}
    \centering
    \begin{tabular}{c|c|c|c|c|c|c}
        \multirow{2}{*}{Env} & \multicolumn{5}{c|}{Mean Fitness} & Human / \\
         & Human & H25 I & H25 F & H0 I & H0 F &  Final H0 \\
        \hline
        G & 7.4  & 5.76 & 8.09 & -0.64 & 6.5 & $113.85\%$ \\
        S & 5.99 & 4.61 & 6.25 & -0.56 & 5.79 & $103.45\%$ \\
        V & 1.24 & 1.02 & 5.83 & -0.06 & 6.52 & $19.02\%$ \\
        \hline
        G(B) & 11.96 & 11.96 & 17.29 & 3.08 & 12.19 & $89.87\%$ \\
        S(B) & 9.2 & 9.2 & 13.89 & 3.05 & 10.68 & $104.31\%$ \\
        V(B) & 3.14 & 9.2 & 10.84 & 2.16 & 13.45 & $26.87\%$ \\
    \end{tabular}
    \label{tab:user_input_quality}
\end{table}

\figurename~\ref{fig:result_stat_plot} contains the curves of mean and best
fitness, QD-score~\cite{pugh2016quality} and coverage at the end of each
iteration, showing the progress of the algorithm running.
\figurename~\ref{fig:result_box_plot} shows MAP-Elites related metrics of the
final archive map, including QD-score, global performance, reliability, and
precision~\cite{mouret2015illuminating}.

\subsection{(RQ1) Affects of human influence during algorithm running}
\label{ssec:rq1}

For mean fitness, the random H0 test condition follows a generally increasing
trend. However, test conditions with human input shows sharp dips in fitness in
the first few iterations: the human designs are relatively high fitness and
early additions to the map decrease the mean. For Ground and Sine, higher
amounts of human input maintain higher average and elite (top 10\% of the
archive by fitness) fitness.  For the Valley, the difficulty of the design task
reduces effectiveness of the human input and the fitness curves largely overlap.

We additionally verified whether human input impacted on the rate of fitness
gain over a run. To do this, we (following the initial dip) recorded the average
iteration that the elite designs of that run achieved a given percentage of the
final mean fitness value.  We compared H25 (typically the overall best
fitness-wise) to H0 on the Ground. \figurename~\ref{tab:fitness_rate} shows that
the H25 case achieved milestones almost twice as fast as the H0 case, indicating
that in addition to providing some benefits to the final design achieved, adding
human input also has the potential to reduce computation by allowing the
algorithm to terminate earlier while still achieving the same quality of
results.  This can be attributed to the subsequent repeated use of
human-designed genetic material as a basis for genetic exploration.  For the
Ground, H25 was able to match the final elite mean fitness of H0 by iteration
1480 on average.  The same test on the Valley shows that H25 fitness values
improved more slowly than H0; in other words, success of the approach relies on
a human's ability to grasp the requirements of the problem.

For the best fitness, the test conditions with human inputs all start at the
same high level on Ground and Sine, as the best human design is guaranteed to be
included in the initial population by the way we select human inputs. And the
curve for H0 starts at a significantly lower point than the rest, and also
converges to a lower value. While for the Valley, the curve of H0 is
indistinguishable to the others. The differences between H0 and other conditions
at the start of the curve in different environments also indicates that the
Ground and Sine are easy environments to the users.

QD-score is a performance measurement for quality diversity algorithm
considering both quality and diversity of the population. For the Ground, all
test conditions have the QD-score curve started roughly at the same point, and
the test conditions with human designs move above H0 and stay up for the whole
time. H25 shows on top of everyone in the late stage. The results of the Sine
show a similar trend, but with narrower gap between H25 and H0. While for the
Valley, H0 gets the best QD-score shortly after a few iterations, and holds the
position for the rest of the run. This implies human designs with bad quality
bring negative effect to QD-score.

Coverage curves show a rapid increase in map coverage at the beginning of the
run, followed by a slower convergence to 100\% coverage.  Lower amounts of human
input achieve higher coverage faster, however 100\% coverage is attained within
400 iterations in all cases, indicating that in our experiment setup, human
input affects only the rising speed of coverage.
\figurename~\ref{tab:coverage_rate} shows the average number of iterations each
test condition takes to reach different percentage of coverage, where numeric
comparisons can be made. \figurename~\ref{tab:coverage_stat_test} shows the
difference in number of generations for each test condition to reach $50\%$ and
$90\%$, with statistical significance verified by the Mann-Whitney U-Test at $p
< 0.05$. This statistically confirms the negative impact of human input on
coverage on all three environments in almost all cases.

Finally, to understand the effect of human input on the rate of coverage, we
recorded the iteration at which each run was able to achieve a given $x\%$
coverage of the archive map, where $x \in \{30, 40, 50, 60, 70, 80,
90\}$. The H0 condition reaches the target coverage in fewer iterations than
H25 and H30 conditions for all environments and all $x$, and H30 performs
worse in terms of coverage than almost all other conditions across all
environments. Despite providing some insight into the dynamics of the
evolutionary process under various conditions, we note that overall coverage
is not affected, reaching $100\%$ in all cases within 400 iterations. In other
words, coverage is not critically affected by the injection of human designs
for the problem considered, although this may differ with differing map size.

\begin{figure}[tb]
\caption{Rate of fitness increase. The columns indicate the iteration at which
the mean fitness of the archive map achieved the listed percentage of the mean
fitness of the final archive.}
\subfloat[Ground]{
    \small
    \begin{tabular}{c|c|c|c|c|c}
        \multirow{2}{*}{Progress} & \multicolumn{5}{c}{Number of iterations} \\
        & H0 & H5 & H15 & H25 & H30 \\
        \hline
         $30\%$ & 101.1 & 51.3 & 25.0 & 25.0 & 25.0 \\
         $40\%$ & 273.7 & 172.2 & 32.5 & 25.0 & 25.0 \\
         $50\%$ & 521.9 & 346.5  & 159.5 & 91.0 & 25.0 \\
         $60\%$ & 772.3 & 544.0  & 346.8 & 358.8 & 122.6 \\
         $70\%$ & 1031.6 & 797.3 & 644.6 & 656.2 & 393.1 \\
         $80\%$ & 1322.7 & 1295.4 & 1200.6 & 1114.0 & 1053.9 \\
         $90\%$ & 1622.8 & 1507.3 & 1517.7 & 1578.1 & 1390.3 \\
    \end{tabular}
} \\
\subfloat[Sine]{
    \small
    \begin{tabular}{c|c|c|c|c|c}
        \multirow{2}{*}{Progress} & \multicolumn{5}{c}{Number of iterations} \\
        & H0 & H5 & H15 & H25 & H30 \\
        \hline
         $30\%$ & 25.0  & 25.0 & 25.0 & 25.0 & 25.0 \\
         $40\%$ & 74.0  & 77.3 & 25.0 & 25.0 & 25.0 \\
         $50\%$ & 274.1 & 278.5 & 82.4 & 108.2 & 25.0 \\
         $60\%$ & 483.2 & 489.2 & 279.9 & 345.9 & 99.0 \\
         $70\%$ & 765.5 & 749.1 & 540.4 & 693.9 & 347.9 \\
         $80\%$ & 922.0 & 1032.3 & 1026.1 & 1135.9 & 1029.2 \\
         $90\%$ & 1512.4 & 1461.0 & 1431.4 & 1569.1 & 1292.0 \\
    \end{tabular}
} \\
\subfloat[Valley]{
    \small
    \begin{tabular}{c|c|c|c|c|c}
        \multirow{2}{*}{Progress} & \multicolumn{5}{c}{Number of iterations} \\
        & H0 & H5 & H15 & H25 & H30 \\
        \hline
         $30\%$ & 64.8  & 25.0 & 92.7 & 98.7 & 110.1 \\
         $40\%$ & 155.8 & 127.8 & 188.8 & 191.3 & 200.8 \\
         $50\%$ & 305.8 & 330.3 & 335.0 & 361.3 & 379.9 \\
         $60\%$ & 508.8 & 569.3 & 611.9 & 647.4 & 671.1 \\
         $70\%$ & 821.8 & 838.7 & 957.0 & 968.3 & 959.2 \\
         $80\%$ & 1225.2 & 1032.1 & 1200.6 & 1293.4 & 1250.2 \\
         $90\%$ & 1626.6 & 1556.1 & 1607.1 & 1568.9 & 1641.6 \\
    \end{tabular}
}
\label{tab:fitness_rate}
\end{figure}

\begin{figure}[tb]
\caption{Rate of coverage. The columns shows the average number of iterations
that take each experiment to reach certain percentage of coverage}
\subfloat[Ground]{
    \begin{tabular}{c|c|c|c|c|c}
        \multirow{2}{*}{Coverage} & \multicolumn{5}{c}{Number of iterations} \\
        & H0 & H5 & H15 & H25 & H30 \\
        \hline
         $30\%$ & 7.0  & 6.9  & 7.3 & 8.3 & 9.2 \\
         $40\%$ & 10.8 & 11.1 & 10.8 & 12.8 & 14.6 \\
         $50\%$ & 15.4 & 15.7 & 15.7 & 18.6 & 21.8 \\
         $60\%$ & 21.5 & 21.4 & 22.4 & 26.9 & 31.1 \\
         $70\%$ & 31.1 & 30.5 & 31.7 & 39.6 & 45.4 \\
         $80\%$ & 45.7 & 44.4 & 47.8 & 57.2 & 67.2 \\
         $90\%$ & 73.4 & 72.0 & 74.1 & 95.9 & 116.7 \\
    \end{tabular}
} \\
\subfloat[Sine]{
    \begin{tabular}{c|c|c|c|c|c}
        \multirow{2}{*}{Coverage} & \multicolumn{5}{c}{Number of iterations} \\
        & H0 & H5 & H15 & H25 & H30 \\
        \hline
         $30\%$ & 6.9  & 7.0 & 6.9 & 8.3 & 11.1 \\
         $40\%$ & 10.5 & 10.9 & 11.2 & 13.6 & 20.3 \\
         $50\%$ & 15.5 & 15.4 & 16.7 & 21.5 & 34.1 \\
         $60\%$ & 21.5 & 21.9 & 24.5 & 33.0 & 53.5 \\
         $70\%$ & 29.8 & 31.1 & 37.1 & 50.1 & 82.3 \\
         $80\%$ & 42.2 & 45.4 & 53.2 & 77.1 & 117.6 \\
         $90\%$ & 64.4 & 70.8 & 81.1 & 121.7 & 174.4 \\
    \end{tabular}
} \\
\subfloat[Valley]{
    \begin{tabular}{c|c|c|c|c|c}
        \multirow{2}{*}{Coverage} & \multicolumn{5}{c}{Number of iterations} \\
        & H0 & H5 & H15 & H25 & H30 \\
        \hline
         $30\%$ & 6.7  & 6.9 & 7.7 & 9.9 & 12.0 \\
         $40\%$ & 10.4 & 10.8 & 12.1 & 15.9 & 19.4 \\
         $50\%$ & 15.1 & 15.6 & 17.7 & 24.6 & 30.6 \\
         $60\%$ & 21.4 & 22.8 & 24.7 & 37.6 & 45.2 \\
         $70\%$ & 30.0 & 32.4 & 34.3 & 54.8 & 64.4 \\
         $80\%$ & 42.3 & 47.5 & 49.7 & 80.4 & 91.1 \\
         $90\%$ & 68.6 & 76.4 & 78.7 & 123.8 & 139.9 \\
    \end{tabular}
}
\label{tab:coverage_rate}
\end{figure}

\subsection{(RQ2) Affects of human inputs on final result}
\label{ssec:rq2}
\figurename~\ref{fig:result_box_plot} shows the statistics of the final result
of experiments. For Ground and Sine, human inputs help to achieve better
QD-score, global performance, reliability and precision than H0. And in most
cases, H25 outperforms other test conditions in all metrics. While for the
Valley, H0 records the best QD-score, precision and reliability. The
Mann-Whitney U-test results for mean archive fitness and best fitness at $P <
0.05$ are shown in \figurename~\ref{tab:fitness_stat_test}. For the Ground and
Sine, the human designs had a positive effect on the mean fitness of the final
archive map, and clearly beats H0 for the best fitness. We however note that for
the Valley, the H0 initial population produced an archive map with a mean
fitness that was statistically significantly higher than almost all other test
conditions. The low mean fitness of the user designs (1.24) compared to the
final mean fitness of the H0 runs (6.52) indicates that users found the Valley
task difficult and biased the search to an unpromising area of the design space.

Results indicate that using only human inputs in the initial population can
hinder the search. For the Ground, H30 condition resulted in a statistically
significantly lower mean fitness in the archive map compared to H25 while for
the Sine, the mean fitness of the two archive maps were approximately the same,
indicating no great benefit from the additional human designs. This implies that
although human intuition was valuable in directing the search to particular
portion of the archive map, there is a benefit to adding random robots to
diversify the population and encourage design exploration. Importantly, the
evolutionary process does not dilute the benefits of using user designs; stark
fitness benefits are observed even after 2000 iterations of evolution with no
additional interaction.  User designs are also frequently replaced in their
niche by evolved designs; evolution uses the provided genetic material to
improve solutions throughout the map.

Given best fitness performance, we conclude that user inputs are beneficial to
evolution in all cases, but more so when the user has a sound grasp of the
problem to be solved.  When this is the case, the users bias the search process
towards promising areas of the design space, as well as providing useful genetic
material for further tinkering by evolution.

\newcommand{\shade}[1]{\cellcolor{black!25}{#1}}
\begin{figure}[]
    \addtolength{\tabcolsep}{-1pt}
    \centering
    \small
    \caption{Comparison on fitness of final archive maps. "$+$" means the
    column result is larger than the row result, while "$-$" indicates the
    opposite. "$\sim$" means difference is less than 0.5\% of the smaller value.
    Shaded cells indicate statistically significant differences confirmed by
    Mann-Whitney U-test at $P<0.05$. ``A'' indicates comparison on mean fitness of
    archive map;  ``B'' indicates comparison on best single fitness of archive map.}
    \subfloat[Ground]{
    \begin{tabular}{c|c c c c}
        \textbf{A} & H5 & H15 & H25 & H30 \\
        \hline
        H0  & +  & \shade{+} & \shade{+} & \shade{+} \\
        H5  &    & +  & \shade{+} & +  \\
        H15 &    &    & \shade{+} & +  \\
        H25 &    &    &    & \shade{-}
    \end{tabular}
    \quad
    \begin{tabular}{c|c c c c}
        \textbf{B} & H5 & H15 & H25 & H30 \\
        \hline
        H0  & \shade{+}  & \shade{+}  & \shade{+}  & \shade{+}  \\
        H5  &    & -  & $\sim$  & +  \\
        H15 &    &    & +  & +  \\
        H25 &    &    &    & +
    \end{tabular}
    }\\
    \subfloat[Sine]{
    \begin{tabular}{c|c c c c}
        \textbf{A} & H5 & H15 & H25 & H30 \\
        \hline
        H0  & $\sim$ & +  & +  & \shade{+} \\
        H5  &    & +  & +  & \shade{+} \\
        H15 &    &    & +  & +  \\
        H25 &    &    &    & $\sim$
    \end{tabular}
    \quad
    \begin{tabular}{c|c c c c}
        \textbf{B} & H5 & H15 & H25 & H30 \\
        \hline
        H0  & + & \shade{+} & \shade{+} & \shade{+} \\
        H5  &    & +  & \shade{+} & \shade{+} \\
        H15 &    &    & \shade{+}  & + \\
        H25 &    &    &    & -
    \end{tabular}
    }\\
    \subfloat[Valley]{
    \begin{tabular}{c|c c c c}
        \textbf{A} & H5 & H15 & H25 & H30 \\
        \hline
        H0  & \shade{-} & +  & \shade{-} & \shade{-} \\
        H5  &    & \shade{+} & -  & -  \\
        H15 &    &    & \shade{-} & \shade{-} \\
        H25 &    &    &    & $\sim$
    \end{tabular}
    \quad
    \begin{tabular}{c|c c c c}
        \textbf{B} & H5 & H15 & H25 & H30 \\
        \hline
        H0  & -  & -  & -  & -  \\
        H5  &    & +  & -  & + \\
        H15 &    &    & -  & + \\
        H25 &    &    &    & +
    \end{tabular}
    }
    \label{tab:fitness_stat_test}
\end{figure}

\begin{figure}[]
    \addtolength{\tabcolsep}{-1pt}
    \small
    \centering
    \caption{Mann-Whitney U test result on coverage progress. For each run,
    the iteration at which that run achieved coverage equal to the
    percentage in the top left corner was computed. The results were compared
    pairwise on the columns vs the rows. The number in the cell indicates the
    difference in iteration number computed for the column minus the row. Shaded
    cells indicate statistically significant differences.}
    \subfloat[Ground]{
    \begin{tabular}{c|c c c c}
        \textbf{$50\%$} & H5 & H15 & H25 & H30 \\
        \hline
        H0  & +0.3  & +0.3 & \shade{+3.2} & \shade{+6.4} \\
        H5  &    & $\sim$  & \shade{+2.9} & \shade{+6.1}  \\
        H15 &    &    & \shade{+2.9} & \shade{+6.1}  \\
        H25 &    &    &    & \shade{+3.2}
    \end{tabular}
    \quad
    \begin{tabular}{c|c c c c}
        \textbf{$90\%$} & H5 & H15 & H25 & H30 \\
        \hline
        H0  & -1.4  & -0.7  & \shade{+22.5}  & \shade{+43.3}  \\
        H5  &    & +2.1  & \shade{+23.9}  & \shade{+44.7}  \\
        H15 &    &    & \shade{+21.8}  & \shade{+42.6}  \\
        H25 &    &    &    & +20.8
    \end{tabular}
    }\\
    \subfloat[Sine]{
    \begin{tabular}{c|c c c c}
        \textbf{$50\%$} & H5 & H15 & H25 & H30 \\
        \hline
        H0  & -0.1 & +1.2  & \shade{+6.0}  & \shade{+18.6} \\
        H5  &    & \shade{+1.3}  & \shade{+6.1}  & \shade{+18.7} \\
        H15 &    &    & \shade{+4.8}  & \shade{+17.4}  \\
        H25 &    &    &    & \shade{+12.6}
    \end{tabular}
    \quad
    \begin{tabular}{c|c c c c}
        \textbf{$90\%$} & H5 & H15 & H25 & H30 \\
        \hline
        H0  & +6.7 & \shade{+17.0} & \shade{+57.6} & \shade{+110.3} \\
        H5  &    & +10.3 & \shade{+50.9} & \shade{+103.6} \\
        H15 &    &    & \shade{+40.6}  & \shade{+93.3} \\
        H25 &    &    &    & \shade{+52.7}
    \end{tabular}
    }\\
    \subfloat[Valley]{
    \begin{tabular}{c|c c c c}
        \textbf{$50\%$} & H5 & H15 & H25 & H30 \\
        \hline
        H0  & +0.5 & \shade{+2.6}  & \shade{+9.5} & \shade{+15.5} \\
        H5  &    & \shade{+2.1} & \shade{+9}  & \shade{+15}  \\
        H15 &    &    & \shade{+6.9} & \shade{+12.9} \\
        H25 &    &    &    & \shade{+6}
    \end{tabular}
    \quad
    \begin{tabular}{c|c c c c}
        \textbf{$90\%$} & H5 & H15 & H25 & H30 \\
        \hline
        H0  & +7.8 & \shade{+10.1} & \shade{+55.2} & \shade{+71.3}  \\
        H5  &    & \shade{+2.3} & \shade{+47.4} & \shade{+63.5} \\
        H15 &    &    & \shade{+45.1} & \shade{+61.2} \\
        H25 &    &    &    & +16.1
    \end{tabular}
    }
    \label{tab:coverage_stat_test}
\end{figure}

\section{Discussion}
\label{sec:conclusion}
In summary, this paper shows the effects of adding human designs to the initial
population of an evolutionary robotics experiment.  Our approach is `set and
forget': it does not require constant/repeated user interaction to achieve its
stated benefits, and is an accessible way to bring intuition into evolutionary
design problems without slowing down evolutionary run time.  It can also be
viewed as the inverse of related approaches wherein computational designs are
suggested to the user for tweaking; here the human does the suggesting and
evolution does the tweaking.

Results demonstrate that when humans have a reasonable understanding of the
problem to be solved, their inputs have beneficial effects on the subsequent
evolutionary process, yielding higher best fitness and higher mean archive
fitness. Evolution biases the initial population into high-performing regions of
the search space (a form of expert knowledge injection), and subsequently
exploits the high-quality genetic material to generate performant designs.  We
note that evolution plays the key designer role in this process; initial human
inputs are not seen in the final populations and are frequently overtaken in
their niches by evolved designs based on their genome parameters. We also see
that the fitness benefits are not simply wiped out by the ongoing genetic
process; instead fitness benefits are lasting and readily observable in the
final evolved populations.  Additionally, archive coverage is not unduly
affected by the use of user designs.  Maximum map coverage is readily achieved
in all cases within 400 iterations, so would only be detrimental in very short
runs that are commonly not used with MAP-Elites. A mixture of human and random
designs (H25) achieves good results in all metrics for Ground and Sine, where
the human intuition is reasonable, suggesting an optimal balance of the two
sources of designs.

We note some limitations with this study.  First, it is dependent on human input
and a cohort of experienced users, which may not be available.  However this
limitation is true of any user-based study. Furthermore, many CAD packages and
robot simulators can readily support this functionality, and many potential
users exist in academia (with access to students) or industry (with access to
peer employees).  The range of applications can therefore be considered to be
relatively large. Finally, results suggest that our approach is highly
beneficial, but is particularly powerful when the human designers have a
reasonable grasp of the problem.  This effect is still to be quantified, but
nevertheless this observation may guide the application of this technique into
other domains.

Future work will focus on assessing the impact of representation on a user's
ability to design, and the interaction of different representations and
evolutionary methods with the process.  We note that Robogami is capable of much
more design diversity than used in this study, so expanding the design space is
a key goal.  Additional studies will explore questions such as: Are there
regions that humans can reach that evolution struggles with, or {\em vice
versa}? Similarly, we wish to exploit the ability to easily fabricate the robots
and move some of the experimentation to reality.

\begin{acks}
This work was conducted at the General Robotics, Automation, Sensing \&
Perception (GRASP) Laboratory at the University of Pennsylvania. The work was
supported in part by the National Science Foundation under Grant \#1845339.
\end{acks}
\balance
\bibliographystyle{ACM-Reference-Format} \bibliography{references.bib}

\end{document}